# A comparative and critical study of EEGNet for fNIRS-driven cognitive load classification

Mehshan Ahmed Khan[a], Houshyar Asadi[a], Li Zhang[b], Mohammad reza Chalak Qazani[c], Ghazal Bargshady[d], Stefanos gkikas[e], Christian arzate[e], Sam Oladazimi[f], Zoran Najdovski[a], Lei Wei[a] and Chee Peng Lim[g]

{mehshan.khan, houshyar.asadi, zoran.najdovski}@deakin.edu.au, Li.Zhang@rhul.ac.uk, mohamadreza.chalakqazani@jcu.edu.au, Ghazal.Bargshady@canberra.edu.au, {stefanos.gkikas, christian.arzate}@jp.honda-ri.com, sam.ozi@lenus-healthcare.de, lei.wei@everlaf.com, cplim@swin.edu.au

[a]*Institute for Intelligent Systems Research and Innovation (IISRI), Deakin University, Geelong Warun Ponds Campus, Victoria 3216, Australia*

[b]*Department of Computer Science, Royal Holloway, University of London, Surrey TW20 0EX, United Kingdom*

[c]*College of Science and Engineering, James Cook University, Townsville, QLD 4814, Australia*

[d]*Faculty of Science and Technology, University of Canberra, Australia*

[e]*Honda research institute (HRI), Japan*

[f]*Hermann-Hesse Str. 17, 72221 Haiterbach, Germany*

[g]*Swinburne University of Technonology, Hawthorn, Victoria, 3122, Australia*

---

**Abstract**

Accurately classifying cognitive load from functional near-infrared spectroscopy (fNIRS) signals remains a significant challenge due to temporal variability, inter-subject differences, and sensitivity to preprocessing choices. This study provides a comprehensive evaluation of EEGNet for fNIRS-based cognitive load classification by systematically examining the effects of temporal segmentation strategies (overlapping vs. non-overlapping), window lengths (10s, 20s, 30s), feature extraction methods (Analysis of Variance (ANOVA), Principal Component Analysis (PCA), Fast Independent Component Analysis (FastICA)), learning rate configurations (fixed and adaptive), and evaluation protocols (random split vs. subject-independent (SI)). Results from random-split experiments show that overlapping segmentation, combined with smaller fixed learning rates (0.01-0.001), yields the highest accuracies, due to temporal redundancy and dense sampling of hemodynamic transitions. However, SI evaluation reveals a substantial drop in accuracy, demonstrating limited generalization to unseen participants. Under SI evaluation, non-overlapping segmentation outperformed overlapping windows, with the best accuracy of 56.11% achieved using PCA features with a 20-second window and a 0.1 learning rate. These findings indicate that eliminating temporal redundancy helps the model learn more robust and generalizable representations of cognitive load across individuals. Although adaptive learning rate strategy improved training stability, it did not surpass the performance of optimally selected fixed learning rates. The study highlights the critical role of segmentation strategy and learning rate selection in improving model generalization and identifies methodological considerations essential for developing reliable, real-time, and SI cognitive load classification systems using fNIRS.

## 1. Introduction

Every year, a significant number of road accidents and collisions occur globally, and a substantial proportion of these incidents are linked to driver distraction or drowsiness [1, 2]. Distraction while driving can arise from a wide range of personal, environmental, and task-related factors. One key contributor is elevated cognitive load, which occurs when drivers engage in secondary tasks such as conversing, using mobile devices, or interacting with in-vehicle technologies [3-5]. Cognitive load reflects the amount of mental effort or working memory resources required to process information and complete a task. When the cognitive demands of a task exceed the available mental resources, an individual's ability to attend to new stimuli or respond to unexpected events diminishes, increasing the risk of errors or delayed reactions. Prolonged engagement in cognitively demanding activities can lead to cognitive fatigue, a state characterized by reduced alertness, slower decision-making, and impaired attention [6, 7]. Drivers experiencing high cognitive load may fail to notice critical traffic signals, overlook pedestrians or obstacles, misjudge distances and speeds, or respond inadequately to sudden hazards. These impairments not only increase the likelihood of collisions but can also reduce a driver's ability to recover from unexpected events, thereby exacerbating the risk of accidents [8, 9]. Understanding and monitoring cognitive load is therefore crucial in safety-critical environments such as driving, where sustained attention and timely decision-making are essential. By identifying periods of high mental demand or fatigue, proactive interventions can be implemented to mitigate risk. Such interventions may include adaptive in-vehicle assistance systems that provide real-time alerts, workload management strategies to distribute cognitive effort more evenly, or partially automated driving support that reduces the driver's cognitive burden during high-demand situations. Effectively managing cognitive load can enhance driver performance, maintain alertness, and ultimately contribute to safer roads.

To reduce the number of road accidents associated with high cognitive load, modern intelligent in-vehicle technologies must be capable of continuously monitoring a driver's mental state and issuing timely alerts when cognitive demands reach unsafe levels. Recent advancements in brain computer interfaces have accelerated progress in this direction by enabling more direct and precise communication between humans and machine systems [10, 11]. When individuals engage in cognitively demanding tasks, their sympathetic nervous system becomes activated, triggering a range of measurable physiological responses. Depending on the intensity of the load, this activation can impact heart rate [12], respiration rate [13], eye-response [14] and blood pressure [15]. These measurable changes provide critical insight into a driver's moment-to-moment cognitive state. Among various sensing technologies, head-worn functional near-infrared spectroscopy (fNIRS) devices have emerged as a promising tool for assessing cognitive load in driving environments [16, 17]. fNIRS enables non-invasive, portable, and relatively robust measurement of cerebral hemodynamics, making it suitable for real-time monitoring of cognitive load [18-20]. With rapid advancements in Artificial Intelligence (AI) across domains such as computer vision [21, 22], pattern recognition [23, 24], and autonomous systems [25, 26], machine learning and deep learning models have become increasingly powerful tools for interpreting complex physiological data. These models are now widely used to translate raw physiological and neurophysiological signals into meaningful indicators of cognitive load. Their ability to learn nonlinear relationships, extract subtle temporal features, and adapt to diverse signal characteristics makes them particularly well suited for real-time cognitive state monitoring in dynamic environments like driving.

This paper provides a comprehensive extension and practical implementation of the concepts introduced in our previous publications [27, 28]. In those earlier studies, our analysis was limited to evaluating EEGNet performance on randomly split datasets using fixed learning rates, and only two feature extraction methods Analysis of Variance (ANOVA) and Principal Component Analysis (PCA) were considered. While those findings offered initial insights into the feasibility of applying EEGNet to fNIRS-based cognitive load classification, they represented only a subset of the broader field of methodology.

In the present work, we significantly expand this scope by incorporating additional feature extraction techniques, including Fast Independent Component Analysis (FastICA) [29, 30], and by systematically examining the impact of variable learning rates through both fixed and adaptive optimization strategies. More importantly, this study introduces a thorough investigation of EEGNet under a subject-independent (SI) evaluation framework, a scenario that we had not previously explored. SI analysis is essential for assessing a model's real-world applicability, as it reflects the performance expected when deploying the system to entirely new users with unseen physiological patterns. Through this expanded methodology, we aim to provide a more complete understanding of EEGNet's robustness, generalizability, and sensitivity to feature selection and training strategies when applied to fNIRS data. The key contributions of this paper are as follows:

1. The study incorporates and compares diverse feature extraction approaches, including FastICA to enable a systematic evaluation of preprocessing strategies and their influence on fNIRS signal and model performance.
2. A comprehensive analysis of fixed, variable, and adaptive learning rate configurations is conducted to assess their effects on training stability, convergence behaviour, and classification accuracy.
3. A thorough SI evaluation of EEGNet is performed to determine its generalization capability, which is essential for real-time cognitive load estimation and deployment on unseen users.
4. Detailed benchmarking is carried out across overlapping and non-overlapping windowing schemes to provide deeper insights into how temporal segmentation choices influence feature representation and overall model performance.

The remainder of this paper is organized as follows. Section 2 presents a review of the relevant literature on cognitive load assessment, fNIRS-based measurement techniques, and deep learning models applied to physiological signals. Section 3 details the complete methodology, including the experimental setup, sensor configuration, driving simulator environment, cognitive load manipulation through the n-back task, data acquisition protocol, preprocessing procedures, feature extraction methods (ANOVA, PCA, and FastICA), normalization techniques, and the implementation and training configuration of the EEGNet model. Section 4 reports and discusses the experimental findings, with emphasis on the effects of temporal segmentation, learning rate strategies, and evaluation protocols. Finally, Section 5 concludes the paper and outlines potential directions for future research.

## 2. Related work

Many prior studies have relied on standardized cognitive tasks to induce measurable and well-controlled levels of mental demand, enabling systematic assessment of cognitive load in laboratory environments. Among the most widely used paradigms is the n-back task, which taxes working memory by requiring participants to continuously compare incoming stimuli with items presented $n$ steps earlier [31-33]. Mental arithmetic tasks, such as serial subtraction or complex equation solving, have also been commonly employed to generate sustained cognitive effort and evaluate workload-related changes in physiological signals [34, 35]. Additionally, the Stroop task is frequently used to induce attentional conflict, and inhibitory control demands by presenting incongruent color–word pairs that require participants to suppress automatic responses [36, 37]. Together, these standardized tasks provide controlled and repeatable methods for eliciting cognitive load, forming the basis for developing and validating predictive models in cognitive workload research.

Cognitive load can be assessed using a combination of subjective and objective measurement techniques, each offering unique insights into an individual's mental demands. Subjective measures rely on self-reporting tools that capture perceived workload, while objective measures examine behavioral or physiological responses. Among subjective assessments, the Paas Cognitive Load Scale (PAAS) [38] remains one of the most widely used tools due to its simplicity and effectiveness in capturing overall mental effort on a single–item rating scale. Another highly established measure is the NASA Task Load Index NASA-TLX [39], a multidimensional questionnaire that evaluates workload across several dimensions, including mental demand, effort, frustration, and temporal pressure. The Subjective Workload Assessment Technique (SWAT) [40] is also frequently employed, offering a structured evaluation of mental load through three key factors: time load, mental effort load, and psychological stress load. Additionally, the Rating Scale Mental Effort (RSME) [41] provides a unidimensional yet sensitive measure of perceived mental effort and is often used in driving, aviation, and ergonomics research. However, self-report measures also come with several important limitations. First, they cannot capture cognitive load frequently or continuously, making them unsuitable for real-time monitoring during dynamic tasks [42, 43]. Their administration is inherently intrusive, as participants must momentarily disengage from the task to provide a rating, which disrupts their ongoing cognitive state and may influence performance [44]. Retrospective self-reports also suffer from recall inaccuracies; participants may reconstruct their mental effort based on memory or assumptions rather than reporting their actual moment-to-moment cognitive load. Additionally, self-reports are vulnerable to subjective biases, including differences in how individuals interpret questions, use numerical scales, or perceive their own mental effort. Factors such as social desirability, misunderstanding of scale anchors, and individual tendencies toward over- or underestimation further reduce the reliability of self-reported cognitive load [45, 46].

These limitations highlight the need for objective, continuous, and non-intrusive measurement approaches, especially in safety-critical environments such as driving where moment-to-moment monitoring of cognitive state is essential. Extensive research offers both theoretical and empirical evidence linking cognitive load to physiological changes, as demonstrated through statistical analysis [47, 48]. As cognitive demands increase, the autonomic and central nervous systems exhibit measurable responses that can be captured through a variety of biosignals. Several physiological and neurophysiological modalities have been used to assess cognitive load. These include Electrodermal Activity (EDA) for sympathetic arousal, skin temperature for stress-related changes [49], and Electrocardiography (ECG) for Heart rate (HR) and Heart rate variability (HRV) [50]. Neurophysiological measures such as Electroencephalogram (EEG) capture cortical activity linked to attention and working memory [51], while eye-tracking, especially pupillometry reflect visual attention and effort [52]. fNIRS has also become a prominent non-invasive method for monitoring hemodynamic responses associated with prefrontal cognitive workload [19].

A wide range of AI algorithms have been employed to analyze cognitive load, particularly for classifying different workload levels from physiological and neurophysiological signals. Traditional machine learning methods, such as Linear Discriminant Analysis (LDA), k-Nearest Neighbors (KNN), and, most commonly, Support Vector Machines (SVM) have been widely adopted due to their simplicity and effectiveness [14, 53, 54]. These approaches rely heavily on two essential steps: feature selection and feature extraction, which determine the quality and discriminability of the input data used for classification. In the fNIRS-based cognitive load studies, researchers have experimented with a variety of hand-crafted features, including changes in Oxygenated Hemoglobin (HbO2), Deoxygenated hemoglobin (HbR), mean HbO/HbR, signal kurtosis and peak amplitudes descriptors [55-57]. While these features can capture meaningful aspects of the hemodynamic response, they require careful manual engineering. This makes performance highly dependent on the researchers' choices and assumptions, potentially limiting model generalizability across tasks or individuals.

Recent advancements in deep learning have addressed several limitations inherent in conventional machine learning approaches by enabling models to learn task-relevant representations directly from raw or minimally processed physiological data. Convolutional Neural Networks (CNNs) have been widely adopted for modalities such as fNIRS, EEG, and ECG due to their ability to automatically extract spatial and temporal patterns through convolutional filters, reducing the need for manual feature engineering [58]. In parallel, Recurrent Neural Networks (RNNs) [16] and their enhanced variants, particularly Long Short-Term Memory (LSTM) [59] networks, have proven highly effective for modeling physiological time-series signals. These architectures capture temporal dependencies and maintain information across longer sequences, making them well suited for characterizing the dynamic fluctuations of cognitive load. Building on these capabilities, Transformer-based models have recently gained attention for cognitive state estimation tasks [60, 61]. Their self-attention mechanisms allow for flexible modeling of long-term temporal relationships without sequential processing constraints, enabling the network to focus on the most informative segments of the signal.

Despite considerable progress in fNIRS-based cognitive load assessment, several critical gaps remain in the current body of research. Many existing studies continue to rely heavily on task performance metrics, such as accuracy, reaction time, or error rates or on subjective self-reports like the NASA-TLX. While useful, these measures provide only post-hoc or coarse estimates of mental workload and fail to capture the participant's true cognitive state in real time. Moreover, the development of robust, real-time models capable of interpreting hemodynamic responses across diverse cognitive tasks remains a significant challenge. A large proportion of existing approaches are task-specific, which limits their generalizability and reduces their suitability for complex, dynamic environments such as driving. Future research must therefore prioritize adaptive fNIRS-based frameworks that can continuously track fluctuations in cognitive workload and support responsive, context-aware intervention systems. Another notable limitation in the literature is the tendency to treat attention or cognitive load as a binary construct for example, “attentive” versus “non-attentive” despite the well-established understanding that cognitive load exists along a continuum and varies considerably across individuals and contexts. This oversimplification restricts the ability of current models to capture the nuanced transitions that occur during real-world multitasking scenarios. Addressing these gaps, the present study contributes to the field by evaluating cognitive load classification using EEGNet, a compact CNN, originally designed for EEG but increasingly applied to hemodynamic data. Importantly, this study conducts an in-depth comparison of random data splits and SI data splits, the latter of which has been rarely examined in fNIRS literature despite its importance for real-world generalizability. By systematically analyzing performance across these evaluation protocols, this work highlights methodological considerations essential for developing reliable, deployable cognitive load assessment systems.

## 3. Research methodology

In this section, we provide a comprehensive description of the dataset used in this study, including details of the experimental setup and data collection procedures. We then introduce the proposed EEGNet model, outlining its architecture and highlighting the key design principles that underpin its structure. In addition, we describe the feature extraction methods applied to the raw fNIRS signals, explaining how these signals were processed and transformed into meaningful representations suitable for classification.

### *3.1. Dataset*

We utilized a dataset that had also been employed in our previous research, where it was used to develop predictive models of cognitive load [27]. The dataset comprised recordings from 38 participants who performed driving tasks within a simulated freeway environment. The simulator was configured with a speed limit of 100 km/h in order to closely replicate real-world highway driving conditions. To further enhance ecological validity, participants were exposed to highly challenging scenarios, including nighttime driving at approximately 1:00 AM combined with adverse weather conditions such as heavy rainfall. These scenarios were intentionally designed to replicate complex real-world driving environments where drivers are often required to maintain vigilance under both environmental and cognitive strain.

During the experimental sessions, participants performed cognitively demanding secondary tasks while simultaneously operating the vehicle in these simulated conditions. The primary secondary task was the auditory n-back, a well-established paradigm for evaluating working memory and mental workload. Participants completed variations of the task (0-back, 1-back, and 2-back), each imposing progressively higher cognitive demands. By incorporating the n-back task into the driving environment, the study was able to systematically manipulate mental workload and examine how increasing cognitive demands influenced both driving performance and physiological responses.

Throughout the experiment, multimodal data streams were collected. fNIRS signals were recorded using a high-density portable neuroimaging system (NIRSIT; OBELAB Inc., Seoul, Korea) to monitor cortical hemodynamic activity, specifically within the prefrontal cortex, a brain region critically involved in attention and executive function. At the same time, simulator-based driving performance metrics were continuously logged, providing complementary behavioral data. Participants were recruited according to clearly defined inclusion criteria to ensure sample consistency and reliability of the data. Eligible participants were required to have prior driving experience to guarantee familiarity with basic vehicle control. Exclusion criteria included any history of neurological, psychiatric, or physical health conditions that could affect cognitive function or interfere with participation. This rigorous selection process was undertaken to reduce variability in baseline cognitive abilities, thereby improving the internal validity of the study and ensuring that observed differences in outcomes were attributable to experimental manipulations rather than individual differences.

### *3.2. Data preprocessing*

The fNIRS dataset employed in this study comprises 204 channels, including 204 $HbO_2$ and 204 HbR signals. To ensure data quality and enhance the reliability of subsequent analyses, several preprocessing steps were applied to the raw signals. These steps included feature selection, normalization, and temporal segmentation, forming a systematic pipeline for dataset preparation. The complete preprocessing workflow is illustrated in Figure 1, which provides an overview of how the raw fNIRS signals were transformed into features suitable for model training and evaluation.

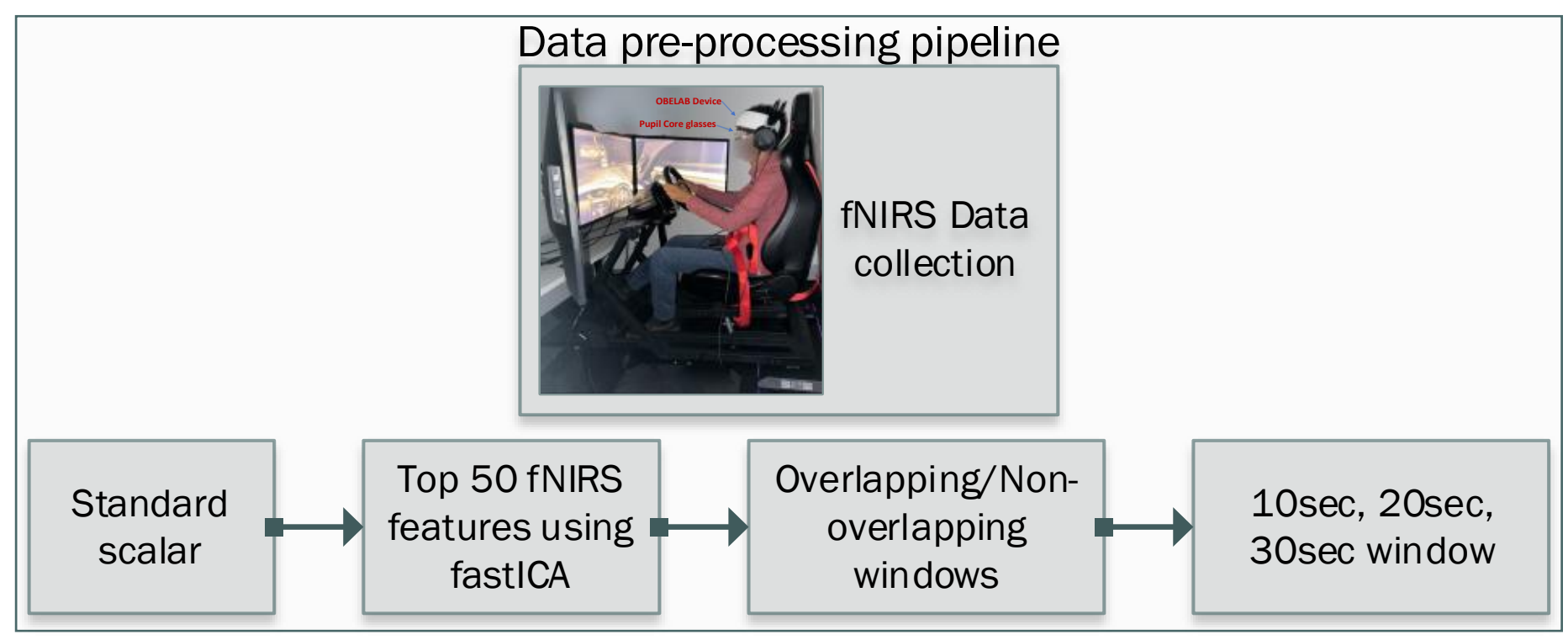

Figure 1: Preprocessing pipeline for fNIRS dataset, illustrating feature extraction, normalization, and segmentation steps prior to model training.

To reduce dimensionality and extract the most informative features, we employed the FastICA algorithm to identify the top fifty features from both $HbO_2$ and HbR signals. ICA is a statistical method designed to decompose multivariate signals into additive, independent non-Gaussian components. In essence, it assumes that observed signals are linear mixtures of latent source signals that are statistically independent. By recovering these independent components, ICA helps reveal hidden structures in high-dimensional data. The FastICA algorithm, as recommended in [62], was adopted due to its efficiency and robustness compared to traditional gradient-based ICA approaches. It relies on a fixed-point iteration scheme, which converges more rapidly and stably than gradient descent methods. This makes it particularly suitable for high-dimensional datasets such as fNIRS. Beyond computational efficiency, FastICA offers two additional advantages. First, it can perform projection pursuit, a method commonly used in exploratory data analysis to uncover interesting data structures and facilitate visualization. Second, it can approximate independent components while identifying low-dimensional projections that capture strongly non-Gaussian distributions [63]. These properties enable FastICA to effectively extract informative spatiotemporal patterns from the fNIRS data, ensuring that the selected $HbO_2$ and HbR features are both meaningful and computationally tractable for subsequent classification tasks.

After feature extraction, we applied normalization to the selected signals in order to standardize feature scales and minimize the influence of varying baseline values across channels. This was performed using the Standard Scaler transformation, shown in Equation (1):

$$X = \frac{X' - \mu}{\theta} \tag{1}$$

Here, $X'$ is the original feature value, $\mu$ represents the mean, and $\theta$ denotes the standard deviation, both computed across the training dataset. Standardization ensures that each channel has zero mean and unit variance, improving numerical stability, accelerating training convergence, and allowing for consistent comparisons across features.

Following standardization, the preprocessed fNIRS data was segmented into temporal windows to capture short-term variations in cognitive load. Temporal segmentation is critical in physiological time-series analysis, as brain activation patterns are inherently dynamic and evolve over time. Two segmentation strategies were investigated:

Overlapping windows, where each new segment shares a portion of its data with the previous one. This sliding approach enhances temporal resolution and enables the detection of transient cognitive state changes. This relationship is formally expressed in Equation (2).

$$X_i = \{x_{i.s}, x_{i.s+1}, \dots, x_{i.s+n-1}\}, i = 0,1,2, \dots \tag{2}$$

where $n$ represents the window length (number of samples per segment), $s$ is the stride or step size between consecutive windows ($s < n$), and $X_i$ is the $i$-th segment. The overlap ensures that consecutive windows contain $n - s$ common samples, thereby capturing finer temporal variations.

In contrast, the non-overlapping window approach divides the signal into consecutive, discrete segments without any overlap. While this method effectively reduces computational complexity and eliminates data redundancy, it may fail to capture subtle or transient transitions in the signal. The formulation of this approach is presented in Equation (3).

$$X_i = \{x_{i.n}, x_{i.n+1}, \dots, x_{(i+1).n-1}\}, i = 0,1,2, \dots \tag{3}$$

Here, each window $X_i$ consists of exactly $n$ samples, with the stride equal to the window length ($s = n$). While this strategy reduces computational overhead and redundancy in the data, it may miss subtle transitions between cognitive states.

To systematically evaluate temporal resolution, we experimented with three window lengths 10 s, 20 s, and 30 s which correspond to 81, 163, and 244 samples at a sampling frequency of 8.138 Hz. This range was chosen based on prior literature on cognitive load estimation with fNIRS, ensuring a balance between capturing sufficient temporal information and maintaining computational efficiency. By comparing these temporal resolutions, we aimed to identify the segmentation strategy and window length that most effectively capture neural dynamics relevant to cognitive load classification.

### *3.3. EEGNet*

EEGNet is a specialized deep learning architecture originally developed for the classification of EEG signals and introduced in 2018 by Lawhern et al. [64]. EEGNet is explicitly optimized for neural signal processing tasks, where datasets are often relatively small and computational efficiency is critical. A key innovation of this architecture lies in its use of depthwise and separable convolutions, which drastically reduce the number of trainable parameters compared to conventional convolutional networks, while still preserving the ability to capture rich spatial and temporal dependencies in the data. This compactness makes EEGNet particularly well-suited for real-time brain–computer interface (BCI) applications and for studies involving resource-constrained environments.

In the present study, we extend the applicability of EEGNet beyond EEG by adopting its architecture for the classification of fNIRS signals, following the implementations of Schirrmeister et al. [65] and Gramfort et al. [66]. This adaptation leverages the architecture's strengths in handling multichannel, time-dependent neural data. A schematic overview of this architecture is presented in Figure 2, which illustrates the layered organization of the model and its flow of information from raw input to final classification. The EEGNet framework is composed of three sequential processing blocks, each designed to progressively transform the raw input into increasingly abstract feature representations. These blocks integrate both spatial and temporal aspects of the signal, enabling the network to effectively distinguish between different levels of cognitive load.

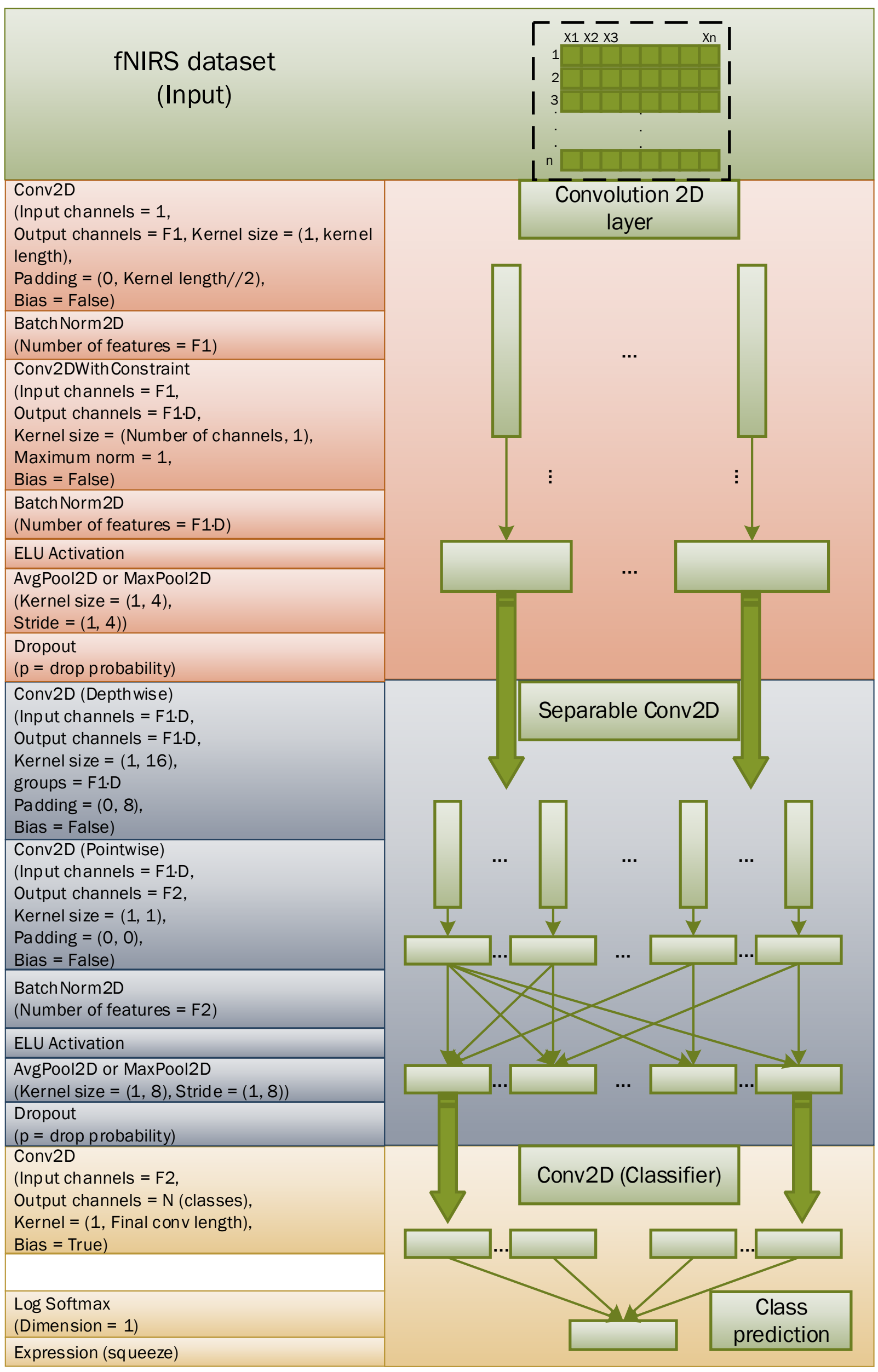

Figure 2: Layered structure of EEGNet illustrating sequential processing blocks and feature extraction for fNIRS signals.

The first block of the proposed EEGNet architecture is designed to extract low-level spatial features from the input signals. It begins with the input layer, followed by a sequence of convolutional operations. Initially, a 2D convolution is applied across both the temporal and spatial dimensions of the input, enabling the network to capture localized features across time and sensor channels. This operation provides the foundation for identifying basic patterns that are relevant to brain activity. Following this, a depthwise convolution is employed. Unlike traditional convolutions that apply a single set of filters across all input channels, depthwise convolution applies a separate filter to each channel individually. This method dramatically reduces the number of trainable parameters, making the model computationally efficient while still maintaining the capacity to capture channel-specific spatial information. To stabilize training and improve convergence, batch normalization is applied after each convolutional step. This ensures that feature distributions remain consistent across layers, thereby enhancing generalization. Importantly, the depthwise convolution also helps to mitigate overfitting.

The second block of EEGNet focuses on capturing spatiotemporal dependencies, which are crucial in understanding how brain signals evolve across time and channels. This block utilizes a separable convolution strategy, which decomposes the learning process into two distinct stages: temporal and spatial integration. The first stage involves a depthwise temporal convolution, where each feature map is filtered independently across time. This operation allows the network to model fine-grained temporal dynamics within each channel, capturing oscillatory or event-related changes that unfold during tasks. The second stage applies a pointwise convolution ($1 \times 1$), which combines the temporally processed signals across channels. By integrating across feature maps, the model learns how spatially distributed signals interact over time, effectively uncovering cross-channel dependencies. This two-step decomposition reduces computational complexity compared to standard convolutions while preserving the model's ability to represent distributed neural activity. For cognitive load classification, this design is especially valuable, as it allows the network to detect subtle variations in both temporal sequences and spatial connectivity patterns, leading to improved sensitivity in distinguishing workload levels.

The third block serves as the classification module of EEGNet, mapping the high-level features learned in the previous layers to the final predictions. The multidimensional feature maps generated by the separable convolution are first flattened into a one-dimensional vector, making them suitable for downstream classification tasks. This flattened representation is then passed to a fully connected (dense) layer, which projects the learned features into a target output space. In this study, the output layer consists of three nodes, each corresponding to one of the defined levels of cognitive load (0-back, 1-back, and 2-back). A SoftMax activation function, as defined in Equation (4), is employed to generate probability distributions over the three classes, enabling the model to output not only the predicted class but also the confidence associated with each prediction.

$$\sigma(\mathrm{z})_{\mathrm{i}} = \frac{e^{z_i}}{\sum_{j=1}^{K} e^{z_j}} \text{ for } i = 1,2, \dots, K \tag{4}$$

where $z_i$ represents the $i^{th}$ logit (raw output) produced by the final dense layer, $K$ denotes the total number of classes (with $K = 3$ in this study), and $\sigma(z)_i$ corresponds to the probability assigned to class $i$.

### *3.4. Data validation framework*

We adopted a 5-fold cross-validation strategy to evaluate the performance and generalizability of the proposed deep learning model. In a standard 5-fold cross-validation, the dataset is partitioned into five equally sized subsets (folds). During each iteration, four folds are used for training while the remaining fold is reserved for testing, and the process is repeated until every fold has served once as the test set. The average performance across the five iterations is then reported, reducing bias associated with a single train-test split.

There are generally two approaches to splitting data in this context, as illustrated in Figure 3. The first approach involves random splitting of the combined dataset into training and testing folds. While this method has been widely reported in earlier studies, it carries significant risks of data leakage and overfitting. Specifically, random splitting may result in data from the same participant appearing in both the training and testing sets. In such cases, the model may inadvertently memorize subject-specific patterns rather than learning generalized features, leading to artificially inflated accuracy.

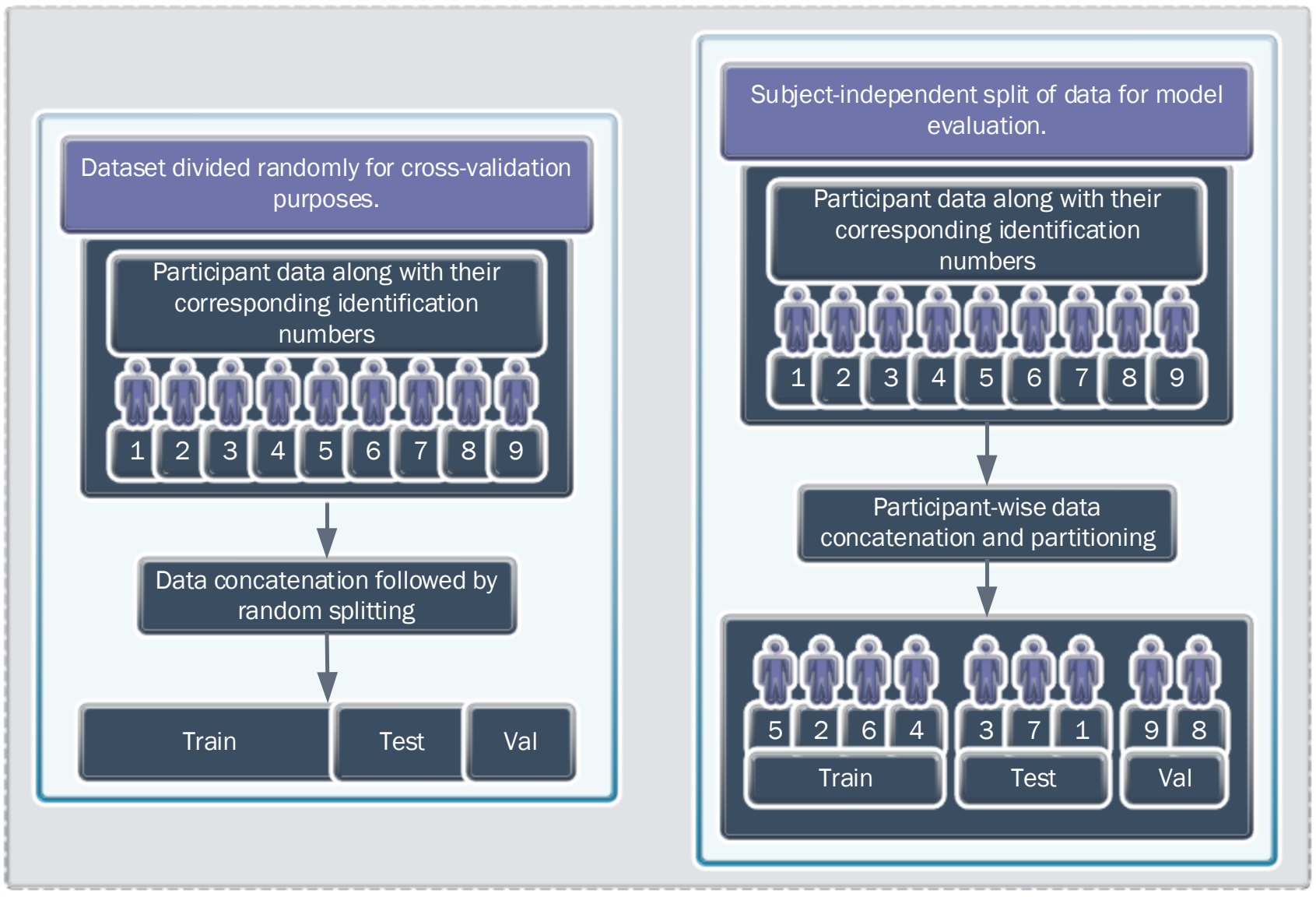

Figure 3:Schematic representation of data partitioning strategies used in the study.

To mitigate this risk, a subject-based splitting strategy is preferred in cognitive neuroscience and physiological signal analysis research. In this approach, data are split at the participant level rather than at the trial level. This means that the training set contains data exclusively from a subset of participants, while the test set includes data from entirely different participants. By doing so, the model is evaluated on its ability to generalize across individuals rather than simply reproducing patterns it has already encountered during training.

This strategy is particularly critical for our dataset, since fNIRS signals exhibit strong inter-individual variability in both hemodynamic response amplitudes and temporal dynamics. Subject-based cross-validation ensures that the model learns patterns associated with cognitive load rather than participant-specific signatures, providing a more realistic estimate of its performance in real-world applications where the model may encounter previously unseen individuals. By comparing both strategies, this study not only highlights the potential pitfalls of random splitting but also demonstrates the importance of subject-based evaluation for building models that can perform reliably in real-world scenarios where data from unseen participants must be classified.

### *3.5. Hyperparameter tuning for classification*

The proposed model was trained using the Adam optimization algorithm, chosen for its ability to adapt the learning rate dynamically during training and thereby accelerate convergence while avoiding poor local minima. The model was optimized with a categorical cross-entropy loss function, which is particularly well-suited for multi-class classification tasks such as distinguishing between different levels of cognitive load. Training was carried out over 200 epochs to allow the network to adequately learn discriminative spatiotemporal patterns from the fNIRS data while maintaining stability in convergence. To ensure consistency and comparability, all core training parameters were fixed across experiments and maintained for each fold of the 5-fold cross-validation. A detailed description of these fixed parameters is presented Table 1.

Table 1: Constant training parameters applied in every fold of the 5-fold cross-validation.

| *Parameter* | **Description / Value** |
|---|---|
| *Dropout strategy* | Standard dropout regularization |
| *Pooling method* | Mean pooling |
| *Activation function* | Exponential Linear Unit (ELU) |
| *Final convolution kernel length* | Automatically determined from data dimensions |
| *Batch size* | 64 samples per batch |
| *Optimizer* | Adam |
| *Loss function* | Categorical cross-entropy |
| *Weight initialization* | Xavier uniform initialization |

To investigate the influence of temporal resolution on model performance, we experimented with three commonly adopted window lengths for cognitive load analysis: 10 s, 20 s, and 30 s. At the sampling rate of 8.138 Hz, these windows correspond to 81, 163, and 244 samples, respectively. Longer windows provide richer temporal information and allow the model to capture extended temporal dependencies in the fNIRS signals; however, they also introduce increased complexity, necessitating deeper convolutional processing to extract both temporal and spatial features effectively. Accordingly, several hyperparameters of the EEGNet architecture were dynamically adjusted to match the input window size. For instance, the temporal convolution kernel length (Kernel 1) in the first Conv2D layer was progressively scaled from 32 for 10-second windows to 128 for 30-second windows, enabling the model to capture a wider temporal context. Similarly, the depthwise convolution kernel (Kernel 2), which focuses on extracting longer-term temporal dependencies, was increased from 8 to 32 across the three window lengths. The number of temporal filters (F1) and pointwise filters (F2) were doubled at each step, thereby expanding the model's representational capacity to account for the higher temporal complexity of longer windows. In contrast, the depth multiplier (D), which governs the degree of spatial filtering, was held constant at 2 across all configurations to maintain a balance between model complexity and training stability. These dynamic hyperparameter adjustments are summarized in Table 2.

Table 2: Adjustments in kernel sizes, filters, and input samples across temporal windows for EEGNet model training.

| *Parameter* | **Description** | **10-second window** | **20-second window** | **30-second window** |
|---|---|---|---|---|
| *Input samples per window* | Number of samples at 8.138 Hz | 81 | 163 | 244 |
| *Dropout rate* | Fraction of units dropped | 0.25 | 0.25 | 0.25 |

| *Temporal kernel size (Kernel 1)* | Length of temporal convolution in first Conv2D layer | 32 | 64 | 128 |
|---|---|---|---|---|
| *Depthwise kernel size (Kernel 2)* | Kernel length for depthwise temporal convolution | 8 | 16 | 32 |
| *Temporal filters (F1)* | Number of temporal filters learned | 8 | 16 | 32 |
| *Depth multiplier (D)* | Scaling factor for spatial filtering | 2 | 2 | 2 |
| *Pointwise filters (F2)* | Number of filters in pointwise convolution (F1 × D) | 16 | 32 | 64 |

To evaluate the impact of learning rate (LR) on classification performance, three different initial values 0.1, 0.01, and 0.001 were systematically tested for each temporal window size (10 s, 20 s, and 30 s). By holding the cross-validation folds constant across all LR configurations, we ensured a fair comparison, eliminating variability that might otherwise arise from differences in data partitioning. This consistency allowed us to directly attribute performance changes to the effect of the learning rate rather than to the data split.

In addition to testing fixed LR values, this study employed a learning rate scheduler to enhance convergence efficiency and training stability. The scheduler continuously monitored the validation loss and reduced the learning rate whenever improvements plateaued. If the validation loss failed to decrease for two consecutive epochs, the scheduler reduced the learning rate by a factor of 0.1. This adaptive adjustment enabled the optimizer to shift from broader exploratory updates to smaller, more fine-grained parameter updates, thereby improving the likelihood of reaching an optimal solution.

**Algorithm 1:** Workflow of EEGNet training with ICA feature selection, multiple initial learning rates, and learning rate scheduler applied across 5-fold cross-validation.

**Input:**

Dataset $D$,
Window sizes $W = \{10s, 20s, 30,$
Sampling frequency $f_s = 8.138$Hz,
Learning rates $LR = \{0.1, 0.01, 0.001\}$ and learning rate schduler
To $K = 50$, features selected via ICA
Number of cross-validation folds $F = 5$,
Number of epochs $E = 200$

**Output:**

EEGNet models $F$ trained for all parameter configurations

**Procedure:**

For each window size $w \in W$:
  Compute the corresponding number of samples $n = w \times f_s$
  Segment the dataset $D$ into sequences of length $n \rightarrow (X, y)$
  Select the top $K$ features from $X$ using ICA $\rightarrow X'$
  Preprocess $X'$ using standard scalar normalization and assemble the dataset $(X', y)$
  For each learning rate $\eta \in LR$ and learning rate scheduler:
    For fold = 1 to $F$ (cross-validation):
      Split data into training and testing sets $\rightarrow \mathcal{D}_{train}, \mathcal{D}_{test}$
      Initialize EEGNet $F$ with window length $n$and configurable parameters $(F_1, F_2, kernel\ size)$
      Configure Adam optimizer with learning rate $n$
      For epoch = 1 to $E$:
        Train $F$ on $\mathcal{D}_{train}$ using cross-entropy loss
        Evaluate performance on $\mathcal{D}_{test}$ and log metrics
      End for
    Store the trained model and results for this window, learning rate, and fold
    End for
  End for
End for
Return all trained EEGNet models

This strategy offers multiple advantages. First, it reduces the risk of the model being trapped in local minima by allowing continued refinement even when progress slows. Second, it helps prevent overfitting by discouraging large parameter updates once the model begins to stabilize. Finally, it reduces the need for manual tuning of the learning rate, ensuring smoother and more efficient training across different configurations and window sizes. The overall training pipeline, including the integration of fixed and dynamic hyperparameters, learning rate experiments, and scheduler adjustments, is summarized in Algorithm 1. This illustration provides a step-by-step overview of how hyperparameter tuning was conducted in this study, highlighting the systematic process adopted to optimize the EEGNet-based model for cognitive load classification using fNIRS data.

## 4. Results and discussion

This section reports the experimental findings from predicting cognitive load using fNIRS data. It covers feature analysis and a comparative evaluation of two data-splitting strategies, random split and SI split, to assess model reliability and generalization across participants.

### *4.1. Feature analysis*

The preprocessing of fNIRS data is a critical step in ensuring that the subsequent classification process is both accurate and computationally efficient. One of the major challenges in fNIRS is its inherently high dimensionality. The device used in this study records data from 204 channels, with each channel providing two hemodynamic signals, $HbO_2$ and HbR. Consequently, the raw dataset consists of a total of 408 features (204 $HbO_2$ + 204 HbR), which poses the risk of redundancy, noise, and overfitting if all features are used directly in the learning stage.

To address this, we began by assessing the quality of individual features using their statistical variance. Mathematically, the variance of the $j^{th}$ feature across $N$ samples can be expressed in Equation (5).

$$\theta_j^2 = \frac{1}{N}\sum_{i=1}^{N}(x_{ij} - \mu_j)^2 \quad (5)$$

where $x_{ij}$ represents the value of feature $j$ for sample $i$, and $\mu_j$ is the mean of feature $j$.

Low-variance features, i.e., those for which $\theta \approx 0$, provide little discriminatory power because they exhibit minimal variation across samples. Such features are often considered uninformative or redundant, and their inclusion may hinder the model by contributing noise rather than useful information. The distribution of feature variances for the entire dataset is shown in Figure 4. As evident from the plot, a large portion of features had very low variance, suggesting limited contribution to classification. These low-variance features were treated as candidates for exclusion.

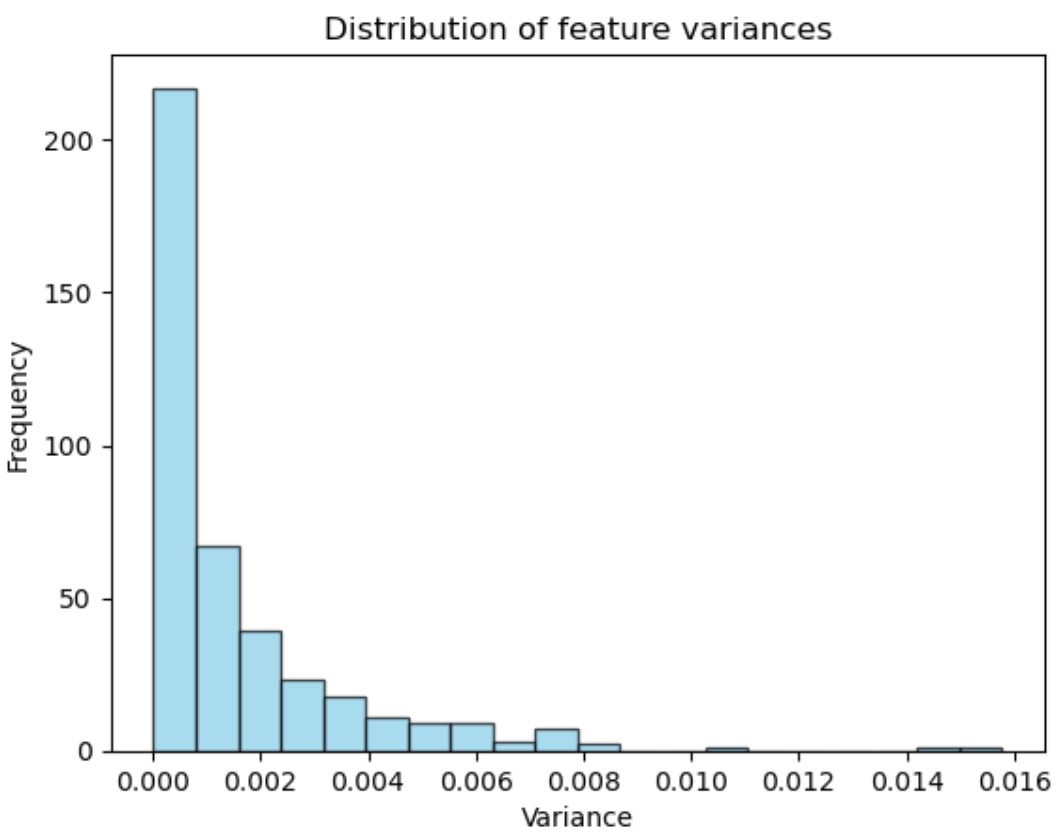

Figure 4: Distribution of feature variances across all fNIRS features. The majority of features exhibit low variance, indicating limited discriminatory power and minimal contribution to classification.

However, variance analysis alone does not always guarantee the identification of the most discriminative features, especially in physiological signals where subtle but independent components may play an important role. To further enhance the feature selection process, we applied FastICA [29, 30] as an additional filtering step. Instead of using the transformed components, FastICA was used to evaluate the contribution of each channel, allowing us to retain only the top fifty most informative $HbO_2$ and HbR features. This approach preserved the interpretability of the original signals while effectively removing less relevant channels and reducing noise in the feature space.

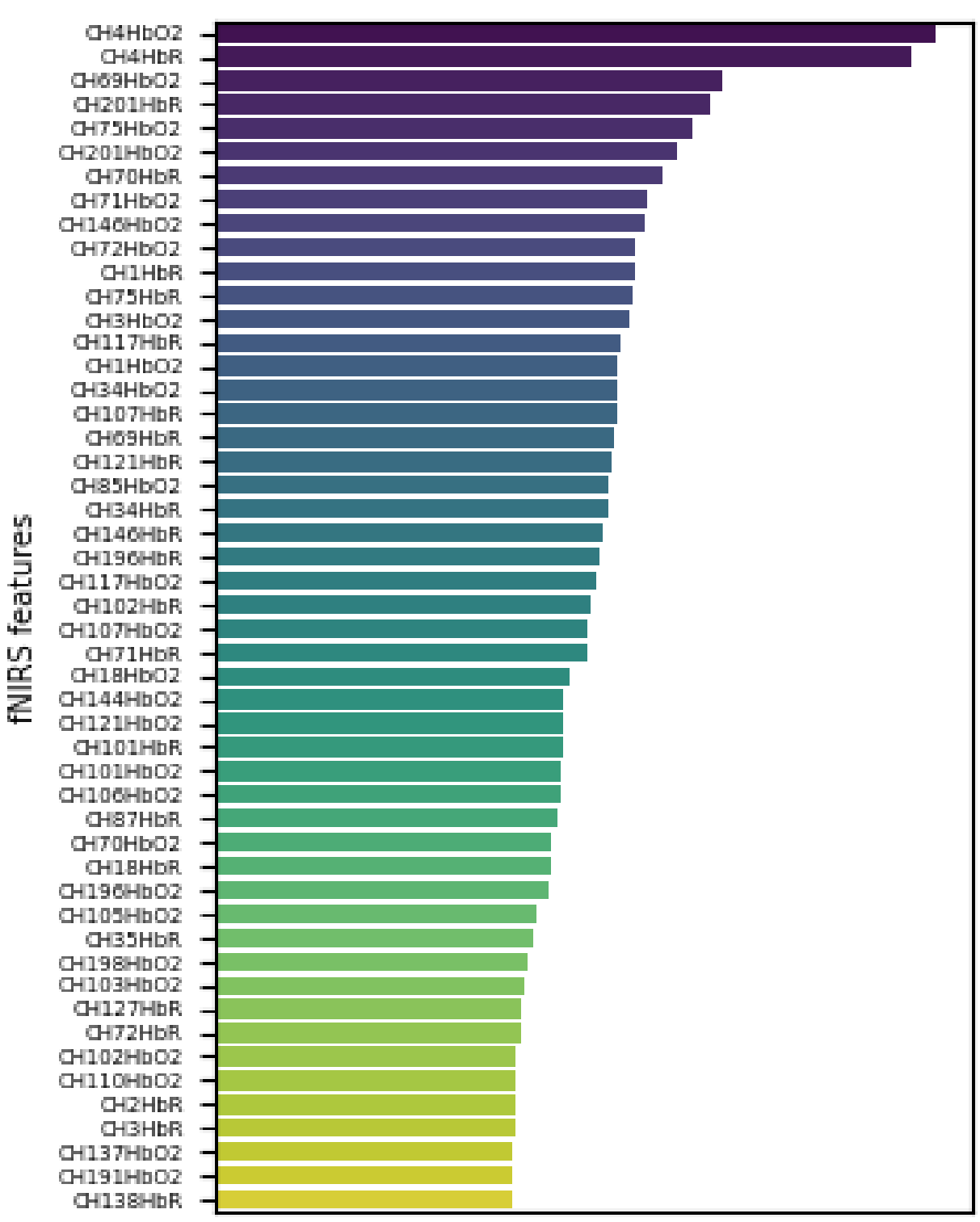

Figure 5: Selected top $HbO_2$ and HbR channels identified using FastICA, along with their relative importance scores. The retained features represent the most informative components for characterizing cognitive load patterns.

The selected features and their importance values, as determined by FastICA, are presented in Figure 5, while their variance distribution is provided in Figure 6. Notably, the retained features exhibited substantially higher variance compared to the unselected ones, confirming their greater ability to capture meaningful variations in brain activation patterns. This outcome demonstrates that FastICA-based selection not only reduced dimensionality but also enhanced the overall feature quality by prioritizing signals with stronger discriminative potential.

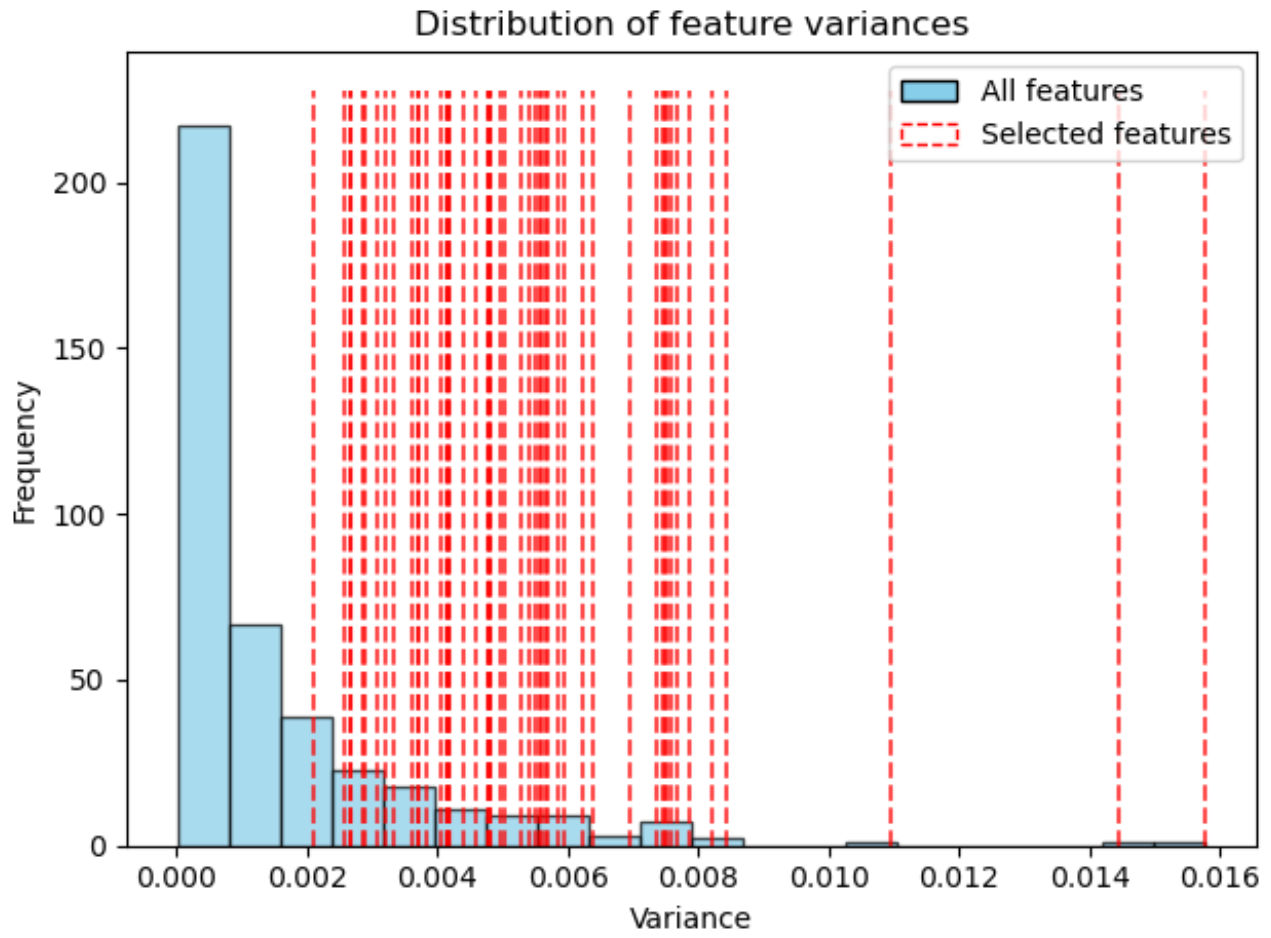

Figure 6: Variance distribution of the FastICA-selected features. Compared to the full dataset (Figure 8), the chosen features exhibit substantially higher variance, confirming their relevance and stronger discriminatory potential for classification.

Beyond analyzing the variance of individual features, the interdependence among the selected features was assessed using the Pearson correlation coefficient. This approach quantifies the linear relationship between pairs of features, providing insight into how much information is shared across channels. The correlation matrix of the FastICA-selected features, illustrated in Figure 7, demonstrates that most features exhibit relatively low correlations with one another. This low level of correlation indicates that the selected features are largely independent, which is useful for DL models because it reduces redundancy and prevents certain features from disproportionately influencing the model. At the same time, some degree of correlation is observed among a few channels, which is expected given that neighboring or functionally related brain regions often show coordinated activity. This moderate correlation does not compromise feature diversity; rather, it highlights that while the features capture distinct patterns, they may also reflect meaningful interactions between brain regions associated with cognitive load. Therefore, the selected channels simultaneously provide independent information and complementary insights into neural activity, which enhances the richness of the input data.

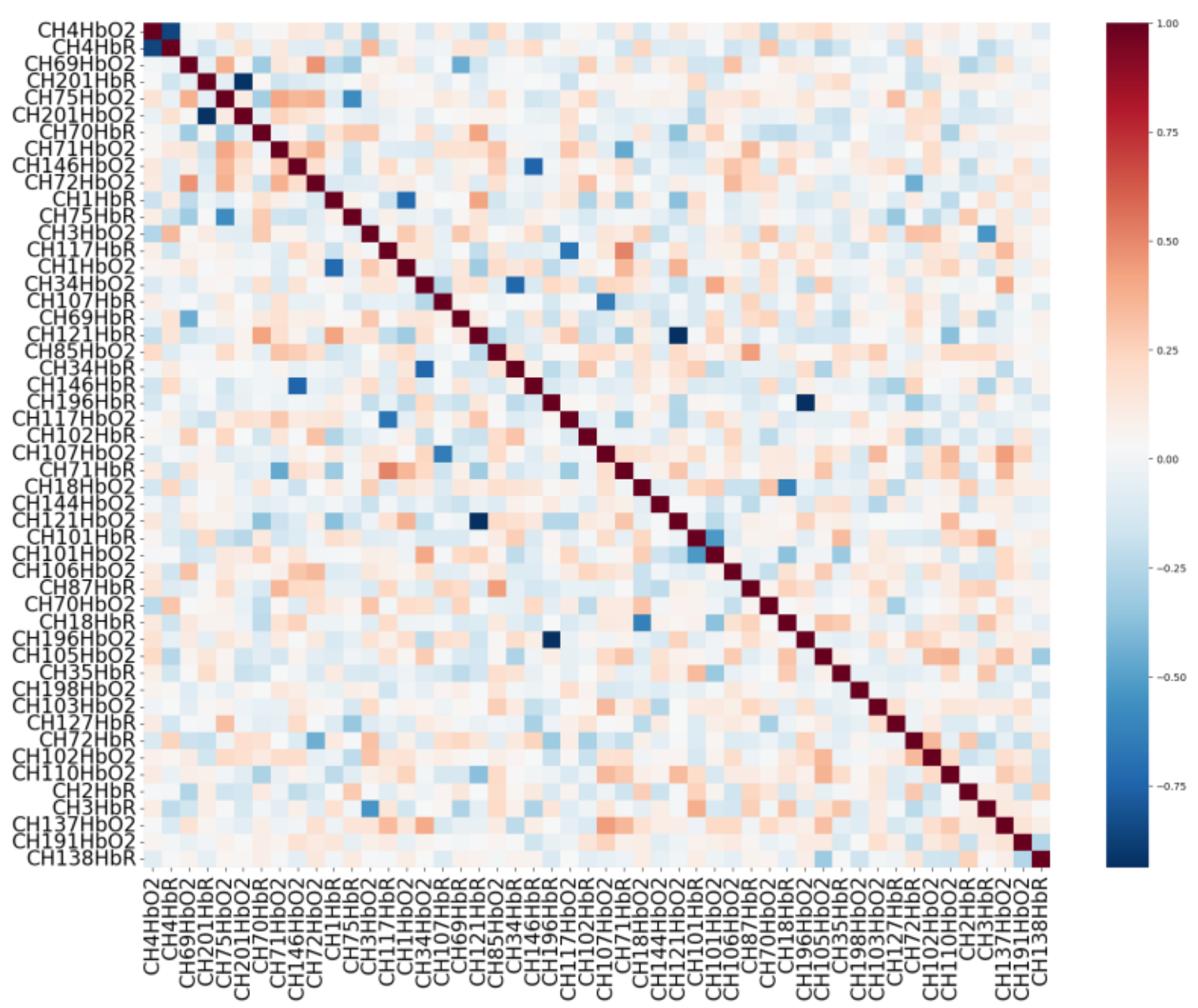

Figure 7: Pearson correlation matrix of FastICA-selected fNIRS features.

*4.2. Model training*

The EEGNet model was trained under multiple experimental configurations to evaluate its robustness and generalization performance. The configurations primarily varied based on window sizes of 10, 20, and 30 seconds, corresponding to different temporal resolutions of the fNIRS data. For each window size, the model architecture and input dimensions were adjusted accordingly to ensure optimal feature representation. Additionally, multiple learning rate settings including fixed and variable learning rate schedules were explored to assess their impact on model convergence and classification accuracy. To examine generalization behavior under different data partitioning strategies, the model was trained using both random split and SI split approaches. Across all configurations and hyperparameter combinations, 5-fold cross-validation was employed. In total, 109 total models were trained to comprehensively assess performance across varying parameter settings.

*4.3. Random data split training*

To evaluate the impact of temporal segmentation on model performance, both overlapping and non-overlapping windowing strategies were applied to the fNIRS signals and analyzed using the EEGNet model. The experiments were performed across three window sizes (81, 163, and 244 samples, corresponding to 10 s, 20 s, and 30 s) and three learning rates (0.1, 0.01, and 0.001), with performance evaluated in terms of accuracy, AUC, recall, precision, and F1-score. The detailed results are summarized in Table 3 (overlapping windows) and Table 4 (non-overlapping windows).

As shown in Table 3, the use of overlapping window segmentation leads to a notable improvement in classification metrics as the learning rate decreases. At a higher learning rate (0.1), the model exhibits moderate accuracy and moderate discriminative capability across all window sizes. However, when the learning rate is reduced to 0.01, there is a significant boost in performance, and at 0.001, the model achieves near-perfect results (≈100%) in all metrics. This trend demonstrates the strong dependence of EEGNet's convergence behaviour on the learning rate. Nonetheless, such exceptionally high performance is likely the result of data redundancy inherent to the overlapping segmentation strategy. Because consecutive overlapping windows share a portion of their data, the model can unintentionally "see" similar patterns in both training and testing samples, leading to data leakage and overfitting. Consequently, while overlapping segmentation improves apparent accuracy, it does not necessarily reflect the model's true generalization capability.

Table 3: Classification performance of EEGNet using FastICA features under overlapping window segmentation and random split configuration. Results indicate that decreasing the learning rate leads to substantial performance improvement, with near-perfect accuracy observed at lr = 0.001, suggesting strong overfitting due to overlapping data samples.

| **Learning rate** | **Window size (samples)** | **Accuracy (Mean ± SD)** | **AUC (Mean ± SD)** | **Recall (Mean ± SD)** | **Precision (Mean ± SD)** | **F1-Score (Mean ± SD)** |
|---|---|---|---|---|---|---|
| **0.1** | 81 (10s) | 0.5686 ± 0.0346 | 0.7436 ± 0.0492 | 0.5686 ± 0.0346 | 0.6063 ± 0.0407 | 0.6063 ± 0.0407 |
| | 163 (20s) | 0.5490 ± 0.0369 | 0.7338 ± 0.0477 | 0.5490 ± 0.0369 | 0.6146 ± 0.0407 | 0.6146 ± 0.0407 |
| | 244 (30s) | 0.5551 ± 0.0291 | 0.7389 ± 0.0343 | 0.5551 ± 0.0291 | 0.5910 ± 0.0208 | 0.5910 ± 0.0208 |
| **0.01** | 81 (10s) | 0.9132 ± 0.0123 | 0.9525 ± 0.0097 | 0.9132 ± 0.0123 | 0.9160 ± 0.0102 | 0.9160 ± 0.0102 |
| | 163 (20s) | 0.9239 ± 0.0130 | 0.9658 ± 0.0106 | 0.9239 ± 0.0130 | 0.9275 ± 0.0108 | 0.9275 ± 0.0108 |
| | 244 (30s) | 0.8937 ± 0.0260 | 0.9509 ± 0.0044 | 0.8937 ± 0.0260 | 0.9029 ± 0.0252 | 0.9029 ± 0.0252 |
| **0.001** | 81 (10s) | 0.9994 ± 0.0003 | 0.9996 ± 0.0002 | 0.9994 ± 0.0003 | 0.9994 ± 0.0003 | 0.9994 ± 0.0003 |
| | **163 (20s)** | **1.0000 ± 0.0001** | **1.0000 ± 0.0001** | **1.0000 ± 0.0001** | **1.0000 ± 0.0001** | **1.0000 ± 0.0001** |
| | **244 (30s)** | **1.0000 ± 0.0000** | **1.0000 ± 0.0000** | **1.0000 ± 0.0000** | **1.0000 ± 0.0000** | **1.0000 ± 0.0000** |

In contrast, Table 4 presents the results obtained from the non-overlapping window segmentation using the same random split approach. This method eliminates redundancy between samples by dividing the signal into distinct, non-repetitive segments, thereby providing a more realistic evaluation of the model's generalization ability. Although the overall performance metrics are lower compared to the overlapping approach, they represent a more reliable and generalizable performance estimate. The results again reveal a consistent pattern, as the learning rate decreases, the model demonstrates better convergence and stability, achieving its best performance at a learning rate of 0.001. The 81-sample window yields the highest accuracy (≈ 97%), indicating that shorter, non-overlapping windows preserve sufficient temporal information for effective classification while maintaining data independence.

Table 4: Classification performance of EEGNet using FastICA features under non-overlapping window segmentation and random split configuration. Compared to overlapping windows, this approach produces lower but more generalizable results, reflecting the reduced redundancy and higher task difficulty.

| **Learning rate** | **Window size (samples)** | **Accuracy (Mean ± SD)** | **AUC (Mean ± SD)** | **Recall (Mean ± SD)** | **Precision (Mean ± SD)** | **F1-Score (Mean ± SD)** |
|---|---|---|---|---|---|---|
| **0.1** | 81 (10s) | 0.6928 ± 0.0290 | 0.8192 ± 0.0371 | 0.6928 ± 0.0290 | 0.7138 ± 0.0344 | 0.7138 ± 0.0344 |
| | 163 (20s) | 0.6650 ± 0.0239 | 0.8247 ± 0.0142 | 0.6650 ± 0.0239 | 0.6692 ± 0.0202 | 0.6692 ± 0.0202 |
| | 244 (30s) | 0.6785 ± 0.0313 | 0.8125 ± 0.0446 | 0.6785 ± 0.0313 | 0.6871 ± 0.0383 | 0.6871 ± 0.0383 |
| **0.01** | 81 (10s) | 0.9550 ± 0.0088 | 0.9816 ± 0.0023 | 0.9550 ± 0.0088 | 0.9559 ± 0.0079 | 0.9559 ± 0.0079 |
| | 163 (20s) | 0.9436 ± 0.0148 | 0.9678 ± 0.0130 | 0.9436 ± 0.0148 | 0.9444 ± 0.0142 | 0.9444 ± 0.0142 |
| | 244 (30s) | 0.9087 ± 0.0220 | 0.9419 ± 0.0393 | 0.9087 ± 0.0220 | 0.9141 ± 0.0179 | 0.9141 ± 0.0179 |

| **0.001** | **81 (10s)** | **0.9726 ± 0.0038** | **0.9888 ± 0.0034** | **0.9726 ± 0.0038** | **0.9728 ± 0.0039** | **0.9728 ± 0.0039** |
|---|---|---|---|---|---|---|
| | 163 (20s) | 0.9335 ± 0.0097 | 0.9664 ± 0.0035 | 0.9335 ± 0.0097 | 0.9350 ± 0.0098 | 0.9350 ± 0.0098 |
| | 244 (30s) | 0.8465 ± 0.0278 | 0.9014 ± 0.0461 | 0.8465 ± 0.0278 | 0.8537 ± 0.0278 | 0.8537 ± 0.0278 |

The 3D surface plot in Figure 8 illustrates the classification performance of EEGNet as a function of window size and learning rate under the overlapping segmentation strategy. A clear interaction pattern is observed, where higher accuracies emerge for intermediate window sizes combined with moderately high learning rates. This trend suggests that overlapping temporal segments provide the model with access to denser, fine-grained temporal information, enabling it to capture subtle transitions in prefrontal hemodynamic activity associated with varying levels of cognitive load. The redundancy inherent in overlapping windows effectively augments the training data, facilitating improved generalization within the overlapping dataset and allowing EEGNet to form more robust spatiotemporal feature representations.

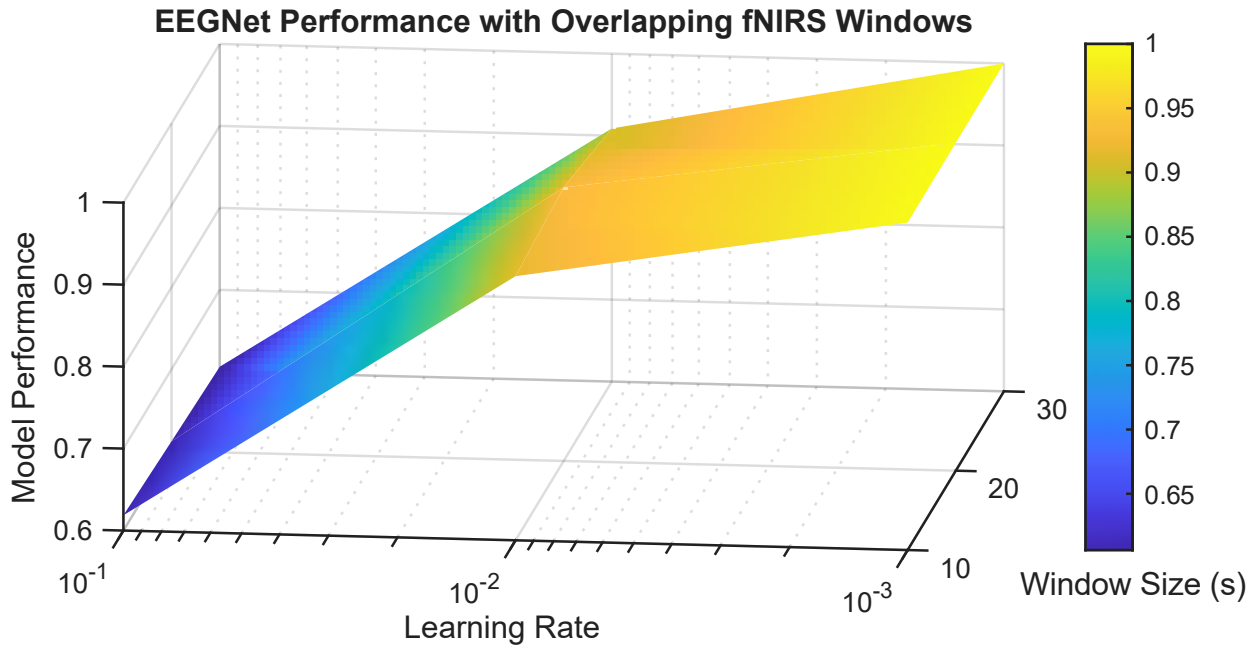

Figure 8: EEGNet model performance (accuracy) across different window sizes and learning rates using overlapping fNIRS signal segments. The 3D surface illustrates how temporal segmentation and learning rate influence classification accuracy, highlighting optimal combinations for real-time cognitive load detection in a driving simulation task.

Similarly, Figure 9 depicts EEGNet's performance using non-overlapping signal segments. In this configuration, overall accuracy is notably reduced, particularly for shorter window lengths and higher learning rates. The absence of overlap leads to a sparser temporal sampling of fNIRS data, limiting the model's ability to capture transient neural dynamics and rapid cognitive fluctuations. Consequently, while non-overlapping segmentation provides a more conservative and independent data structure, reducing redundancy and the risk of overfitting, it does so at the cost of reduced temporal continuity and representational richness.

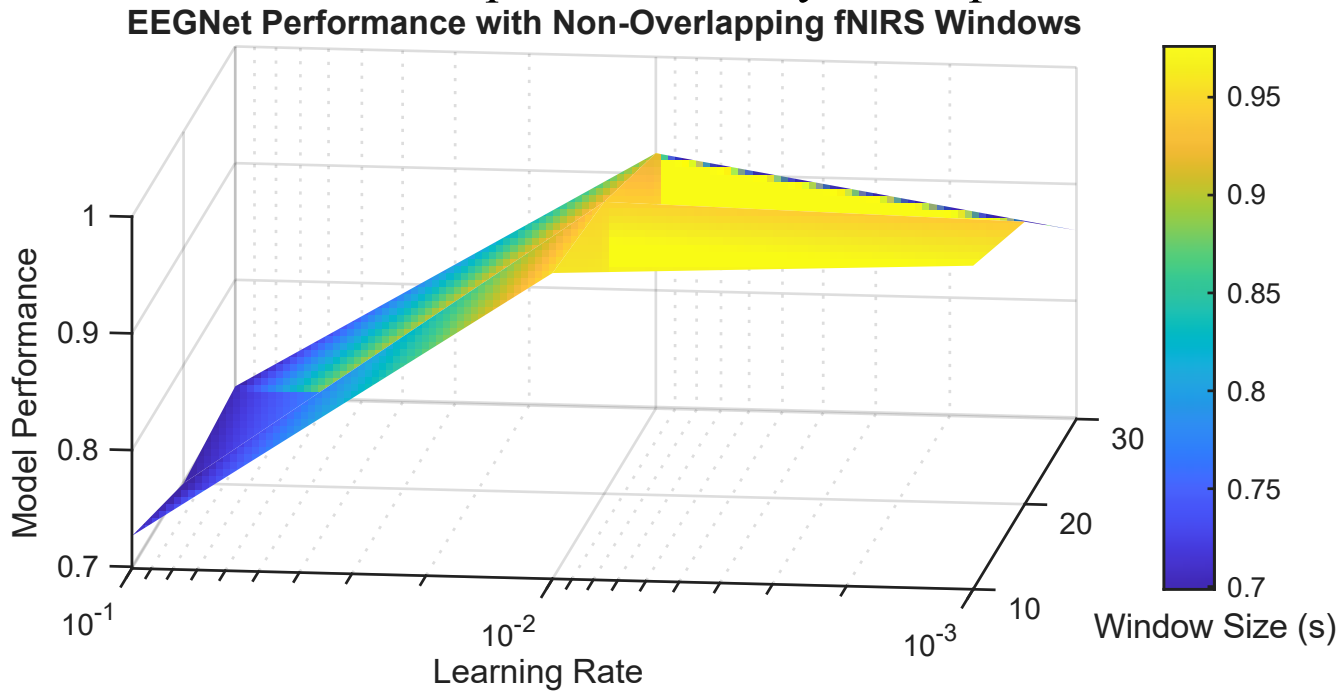

Figure 9: EEGNet model performance (accuracy) across different window sizes and learning rates using non-overlapping fNIRS signal segments. The surface plot shows the effect of window segmentation on model performance, emphasizing the differences compared to overlapping segmentation strategies for cognitive load classification.

When comparing the two segmentation strategies, overlapping windows consistently outperform non-overlapping ones across most parameter combinations, particularly for smaller window sizes and intermediate learning rates. This performance gap underscores the importance of temporal context in modeling cognitive load from fNIRS signals. Overlapping segmentation enhances temporal resolution and statistical power, whereas non-overlapping segmentation prioritizes data independence and generalizability. Together, these findings highlight a critical methodological trade-off between maximizing classification accuracy and preserving temporal distinctiveness in neuroimaging-based cognitive load analysis.

Figure 10 illustrates the relative improvement in EEGNet model performance when using overlapping versus non-overlapping fNIRS signal segments. The 3D bar plot presents the normalized difference in accuracy across multiple window sizes (10, 20, and 30 seconds) and learning rates (0.1, 0.01, and 0.001), providing a comprehensive view of how segmentation strategy interacts with model hyperparameters. Overall, overlapping windows consistently yield higher performance compared to non-overlapping ones, with the greatest improvements observed at shorter window sizes. This finding suggests that overlapping temporal segments offer a richer and more continuous temporal representation of the fNIRS signals, enabling the model to better capture dynamic fluctuations in cognitive load. The extent of improvement also depends on the learning rate—intermediate values (e.g., 0.01) typically produce the most pronounced gains, likely reflecting an optimal balance between convergence speed and stability during training. In contrast, smaller learning rates achieve less benefit from overlap, possibly due to already stable optimization dynamics. The influence of window size further reveals that smaller windows (10 s) benefit substantially from overlapping segmentation, whereas larger windows (30 s) show diminishing returns, as they inherently encompass sufficient temporal context.

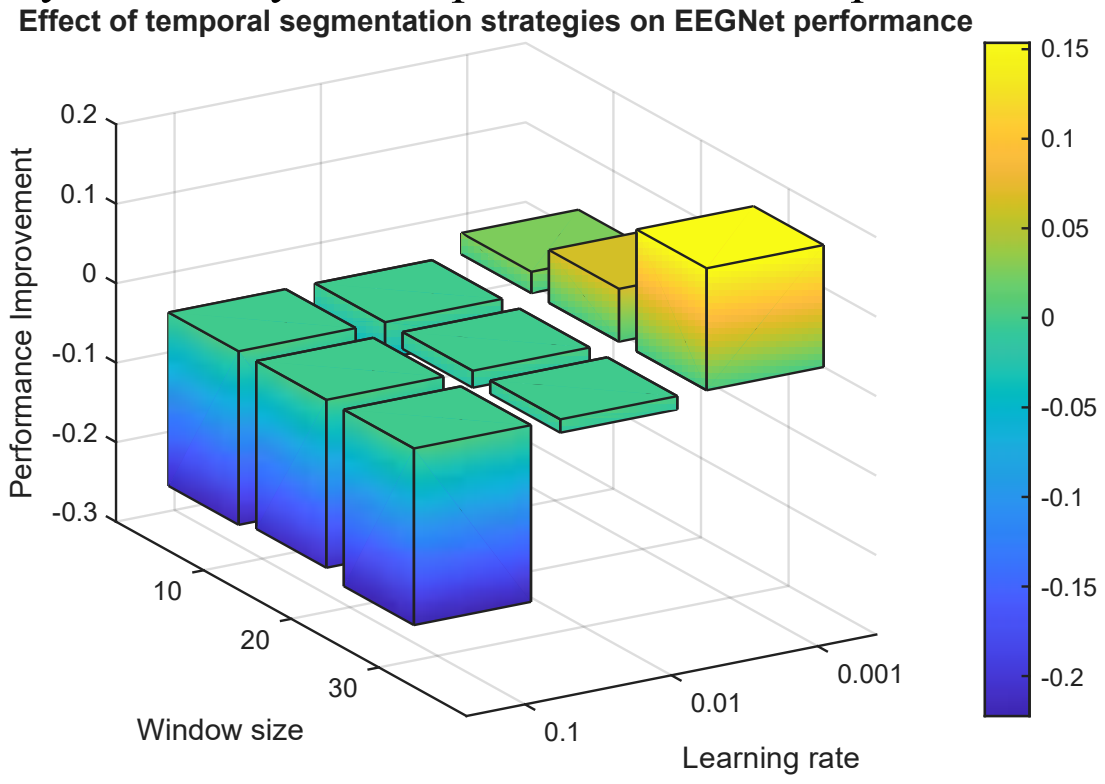

Figure 10: Comparative analysis of EEGNet model performance improvement using overlapping versus non-overlapping fNIRS signal segments. The 3D bar plot represents the relative performance improvement across different window sizes (10, 20, 30 s) and learning rates (0.1, 0.01, 0.001), highlighting how temporal segmentation influences accuracy in cognitive load classification during a driving simulation task.

In our previous studies, we have employed ANOVA [27] and PCA [28] feature extraction techniques in conjunction with the EEGNet model, using identical model configurations and experimental parameters as in the present work. As illustrated in Figure 11a, Figure 11b and Figure 11c, the comparative results indicate that the average model performance across the three learning rates (0.1, 0.01, and 0.001) and window sizes (81, 163, and 244 samples) under the overlapping segmentation strategy remains largely consistent across all feature extraction approaches. Although minor fluctuations in accuracy, AUC, recall, precision, and F1-score can be observed, these variations are relatively small and do not significantly influence the overall performance trend. This consistency suggests that the EEGNet model demonstrates a stable and robust learning capability across different feature representations derived from fNIRS data. Specifically, regardless of whether ANOVA, PCA, or FastICA is used to extract features, the model successfully captures the underlying temporal and spatial characteristics necessary for effective classification of cognitive load levels.

This observation also implies that the feature extraction method, while contributing to slight differences in signal representation, may not substantially alter the discriminative information embedded within the prefrontal fNIRS signals. Instead, EEGNet's convolutional and depthwise separable filtering structure appears to automatically learn task-relevant features from the input representation, minimizing the dependency on specific preprocessing or dimensionality reduction techniques. Consequently, the similar performance obtained across ANOVA, PCA, and FastICA-based feature inputs reinforces the adaptability of EEGNet to various data transformations, making it a reliable framework for multimodal neuroimaging analysis.

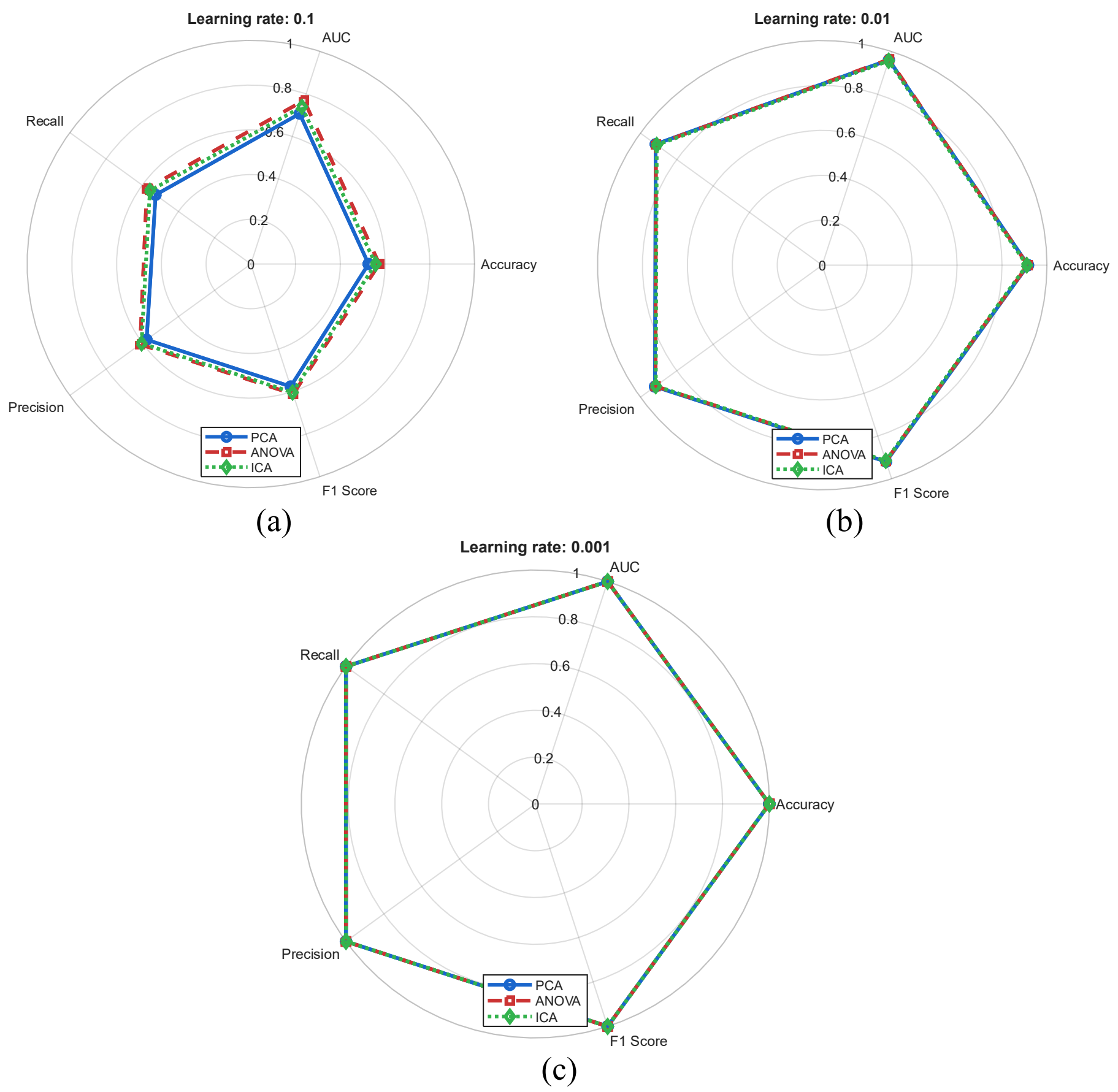

Figure 11: EEGNet classification performance across different feature extraction methods using overlapping window segmentation of fNIRS data. (a) ANOVA [27] features, (b) PCA [28] features, and (c) FastICA-features are evaluated across multiple learning rates (0.1, 0.01, 0.001). The plots illustrate consistent performance patterns across feature extraction techniques, with overlapping segmentation yielding higher accuracy and smoother performance transitions across hyperparameter combinations.

Comparable patterns were also observed in the non-overlapping segmentation experiments, as shown in Figure 12a, Figure 12b and Figure 12c where model performance across the three feature extraction approaches followed a similar trend with only marginal variations in the evaluation metrics. However, in all cases, the overlapping window segmentation consistently yielded superior results compared to non-overlapping segmentation. This performance gap can be attributed to the temporal redundancy introduced by overlapping segments, which allows the model to access partially shared temporal information across consecutive samples. Such redundancy enhances temporal continuity and provides the network with a richer representation of dynamic hemodynamic changes linked to cognitive load fluctuations. In contrast, non-overlapping segmentation, while reducing redundancy and ensuring data independence, limits the amount of temporal information per segment and thus constrains the model's ability to detect rapid variations in neural activity.

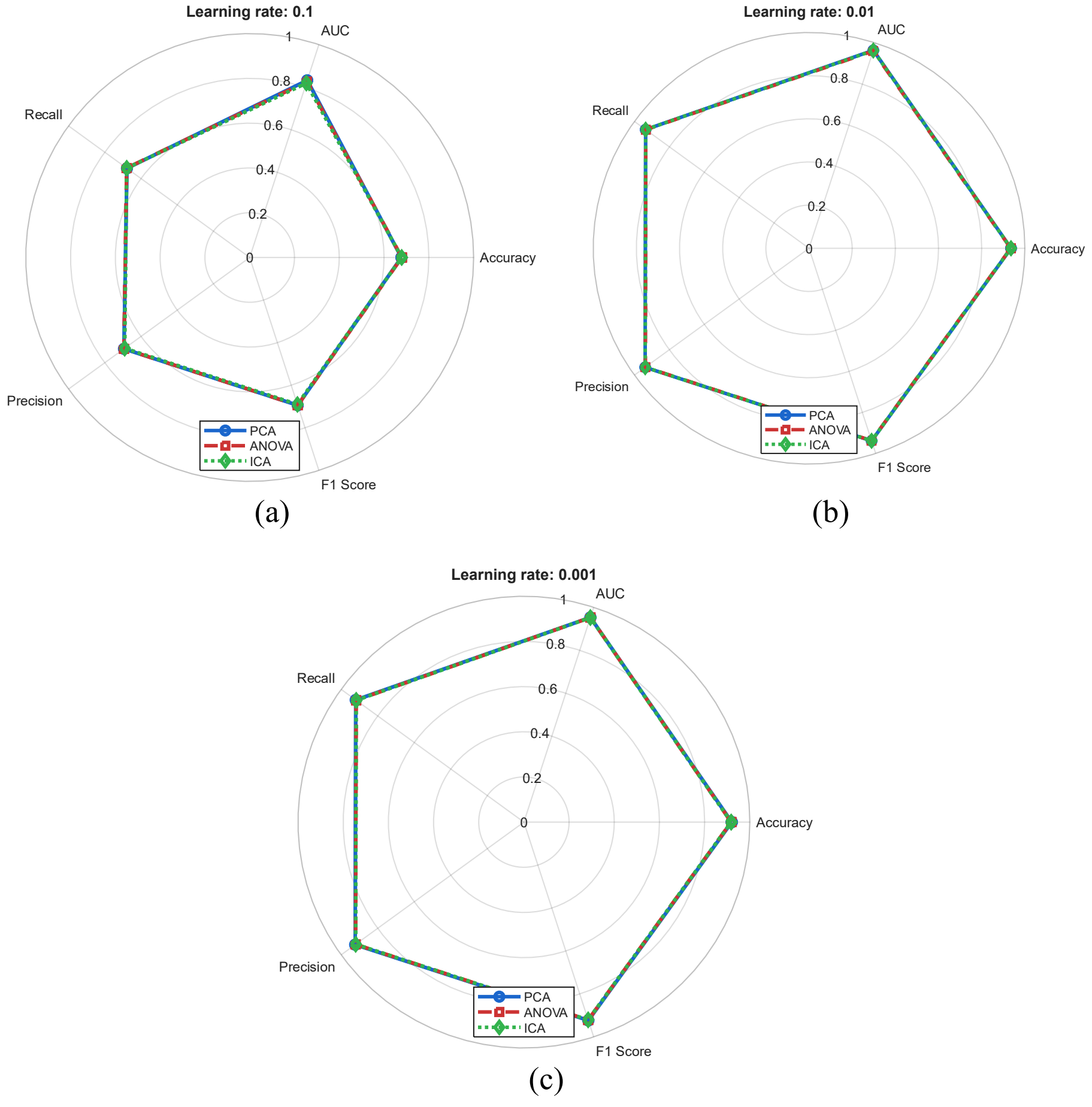

Figure 12: EEGNet classification performance across different feature extraction methods using non-overlapping window segmentation of fNIRS data. (a) ANOVA [27] features, (b) PCA [28] features, and (c) FastICA-based features are evaluated across multiple learning rates (0.1, 0.01, 0.001). While overall accuracies are slightly reduced compared to overlapping segmentation, the trends remain consistent across feature extraction methods, highlighting the stability of EEGNet and the trade-off between temporal continuity and data independence in cognitive load classification.

In this study, we have also incorporated and evaluated an automatic learning rate adaptation technique, which was not explored in our previous work. Specifically, we employed the Reduce Learning Rate on Plateau (RLRP) method [67] strategy, a widely used optimization method that dynamically adjusts the learning rate during training to enhance model convergence and stability. As the name suggests, this method monitors the validation loss and automatically reduces the learning rate once the model's performance ceases to improve, allowing for finer optimization as training progresses. In our implementation, we initialized the learning rate at 0.1 and set a patience value of 2, meaning that if the validation loss does not improve for two consecutive epochs, the learning rate is reduced by a factor of 0.1. This adaptive approach helps prevent premature convergence and mitigates oscillations in the loss function, especially during the later stages of training. Figure 13a presents the training loss behaviour for models trained with ANOVA features under both overlapping and non-overlapping segmentation, showing how the Reduce Learning Rate on Plateau strategy lowers the learning rate whenever the loss reaches a plateau, enabling smoother and more stable convergence. Figure 13b shows a similar trend for PCA features, where the adaptive reduction in learning rate helps the model escape stagnant regions of the loss surface and continue optimizing effectively. Figure 13c illustrates the loss curves for FastICA features, again demonstrating consistent learning rate adjustments that prevent oscillations and support gradual refinement of model parameters. Across all three feature extraction methods, the adaptive behaviour clearly stabilizes the training process and promotes more robust performance by allowing the model to fine-tune its weights when further improvements become difficult to obtain with the initial learning rate.

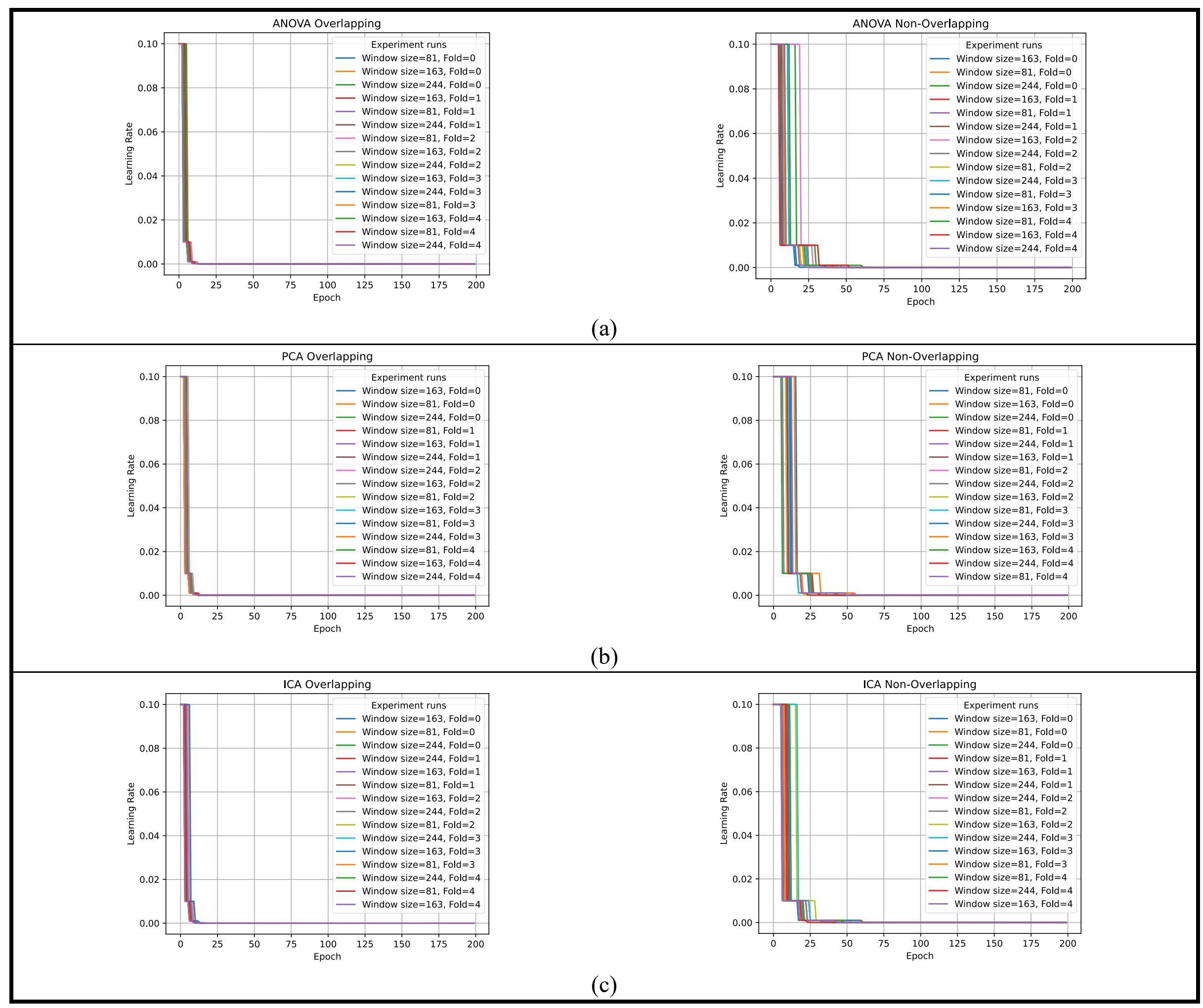

Figure 13: Training loss behaviour for EEGNet using (a) ANOVA, (b) PCA, and (c) FastICA features with overlapping and non-overlapping segmentation under the random data split. The curves show how the RLRP strategy adaptively lowers the learning rate when validation loss plateaus, resulting in more stable convergence across all feature extraction methods.

Unlike fixed learning rates used in earlier experiments, this method dynamically adjusts the learning rate during training, reducing it by a factor of 0.1 when the validation loss fails to improve for two consecutive epochs. As shown in Table 5, overlapping window segmentation produced exceptionally high performance across all feature extraction methods and window sizes, with accuracies exceeding 99.7% and AUC values approaching 1.0. These results demonstrate that automatic learning rate adaptation enables the model to achieve near-perfect convergence while maintaining consistent precision, recall, and F1-scores across different feature extraction pipelines. The minimal variation across ANOVA, PCA, and FastICA suggests that the adaptive learning rate mechanism allows EEGNet to learn efficiently regardless of the input feature space, effectively compensating for minor differences in feature representation. However, this extremely high performance also reflects the inherent data redundancy caused by overlapping windows, which may introduce partial data leakage and inflate evaluation metrics.

Table 5: Classification performance of the EEGNet model using ANOVA, PCA, and FastICA feature extraction methods respectively, under overlapping window segmentation with automatic learning rate adaptation.

| **Sample size** | **Feature extraction methods** | **Accuracy (Mean ± SD)** | **AUC (Mean ± SD)** | **Recall (Mean ± SD)** | **Precision (Mean ± SD)** | **F1-Score (Mean ± SD)** |
|---|---|---|---|---|---|---|
| **81 (10s)** | ANOVA | 0.9977 ± 0.0008 | 0.9992 ± 0.0004 | 0.9977 ± 0.0008 | 0.9977 ± 0.0008 | 0.9977 ± 0.0008 |
| | PCA | 0.9977 ± 0.0009 | 0.9992 ± 0.0003 | 0.9977 ± 0.0009 | 0.9978 ± 0.0009 | 0.9978 ± 0.0009 |
| | FastICA | 0.9982 ± 0.0005 | 0.9993 ± 0.0004 | 0.9982 ± 0.0005 | 0.9982 ± 0.0005 | 0.9982 ± 0.0005 |
| **163 (20s)** | ANOVA | 0.9996 ± 0.0002 | 0.9997 ± 0.0001 | 0.9996 ± 0.0002 | 0.9996 ± 0.0002 | 0.9996 ± 0.0002 |
| | PCA | 0.9990 ± 0.0012 | 0.9996 ± 0.0003 | 0.9990 ± 0.0012 | 0.9990 ± 0.0012 | 0.9990 ± 0.0012 |

| | FastICA | 0.9996 ± 0.0003 | 0.9998 ± 0.0001 | 0.9996 ± 0.0003 | 0.9996 ± 0.0003 | 0.9996 ± 0.0003 |
|---|---|---|---|---|---|---|
| **244 (30s)** | ANOVA | 0.9996 ± 0.0003 | 0.9997 ± 0.0002 | 0.9996 ± 0.0003 | 0.9996 ± 0.0003 | 0.9996 ± 0.0003 |
| | PCA | 0.9997 ± 0.0005 | 0.9999 ± 0.0002 | 0.9997 ± 0.0005 | 0.9997 ± 0.0005 | 0.9997 ± 0.0005 |
| | FastICA | 0.9998 ± 0.0002 | 1.0000 ± 0.0001 | 0.9998 ± 0.0002 | 0.9998 ± 0.0002 | 0.9998 ± 0.0002 |

In contrast, Table 6 presents results for non-overlapping windows, which offer a more realistic evaluation of model generalization. While the overall accuracies are slightly lower compared to overlapping segmentation, the results remain strong, with accuracies ranging from approximately 85% to 96% and AUC values between 0.91 and 0.98 across different feature extraction methods. These findings indicate that the learning rate scheduler effectively stabilizes learning and prevents premature convergence, even under stricter data independence conditions. Notably, shorter window sizes (10 s, 81 samples) yield the highest performance, suggesting that finer temporal resolution captures essential cognitive state variations while benefiting from adaptive learning rate optimization.

Table 6: Classification performance of EEGNet using ANOVA, PCA, and FastICA features respectively, under non-overlapping window segmentation with automatic learning rate adaptation. While overall accuracy is lower than overlapping segmentation, results demonstrate strong generalization and improved robustness across all feature extraction methods.

| **Sample size** | **Feature extraction methods** | **Accuracy (Mean ± SD)** | **AUC (Mean ± SD)** | **Recall (Mean ± SD)** | **Precision (Mean ± SD)** | **F1-Score (Mean ± SD)** |
|---|---|---|---|---|---|---|
| **81 (10s)** | ANOVA | 0.9579 ± 0.0125 | 0.9784 ± 0.0098 | 0.9579 ± 0.0125 | 0.9580 ± 0.0127 | 0.9580 ± 0.0127 |
| | PCA | 0.9638 ± 0.0137 | 0.9815 ± 0.0076 | 0.9638 ± 0.0137 | 0.9640 ± 0.0138 | 0.9640 ± 0.0138 |
| | FastICA | 0.9638 ± 0.0054 | 0.9830 ± 0.0041 | 0.9638 ± 0.0054 | 0.9639 ± 0.0054 | 0.9639 ± 0.0054 |
| **163 (20s)** | ANOVA | 0.9040 ± 0.0221 | 0.9583 ± 0.0076 | 0.9040 ± 0.0221 | 0.9057 ± 0.0206 | 0.9057 ± 0.0206 |
| | PCA | 0.9024 ± 0.0306 | 0.9580 ± 0.0072 | 0.9024 ± 0.0306 | 0.9050 ± 0.0295 | 0.9050 ± 0.0295 |
| | FastICA | 0.9167 ± 0.0208 | 0.9651 ± 0.0124 | 0.9167 ± 0.0208 | 0.9182 ± 0.0203 | 0.9182 ± 0.0203 |
| **244 (30s)** | ANOVA | 0.8505 ± 0.0240 | 0.9225 ± 0.0345 | 0.8505 ± 0.0240 | 0.8592 ± 0.0187 | 0.8592 ± 0.0187 |
| | PCA | 0.8518 ± 0.0317 | 0.9193 ± 0.0354 | 0.8518 ± 0.0317 | 0.8618 ± 0.0245 | 0.8618 ± 0.0245 |
| | FastICA | 0.8518 ± 0.0268 | 0.9187 ± 0.0354 | 0.8518 ± 0.0268 | 0.8615 ± 0.0206 | 0.8615 ± 0.0206 |

These results emphasize the strong influence of temporal segmentation and learning rate choice on EEGNet performance for fNIRS-based cognitive load classification. While overlapping segmentation boosts apparent accuracy due to temporal redundancy, non-overlapping segmentation provides more reliable generalization. Notably, the fixed learning rate of 0.001 consistently outperformed the automatic learning rate scheduler, suggesting that a well-chosen constant learning rate can yield more effective and stable optimization than adaptive scheduling in this experimental context.

*4.4. Subject-Independent Split training*

We have also employed a SI evaluation approach across the various feature extraction methods (ANOVA, PCA, and FastICA), using different learning rates and window sizes in the same manner as the model was evaluated under the random split condition. However, instead of randomly splitting the data across all samples, we applied a 5-fold cross-validation scheme based on subjects, where each fold represents data from distinct participants. This ensures that the model is trained and validated on mutually exclusive subject sets, thereby testing its ability to generalize across individuals. The SI approach is widely adopted in the literature as a robust method for assessing a model's generalization capability and suitability for real-time inference [68-70]. Unlike subject-dependent models that rely on data from the same individuals for both training and testing, this setup evaluates the model's performance on entirely unseen participants [69]. Consequently, it provides a more realistic representation of practical, real-world scenarios such as real-time cognitive load monitoring in driving environments, where the model must reliably classify cognitive states for new users without requiring additional calibration or personalized retraining.

Table 7 presents the results of the SI evaluation for overlapping window segments using the three-feature extraction methods ANOVA, PCA, and FastICA across different learning rates (0.1, 0.01, and 0.001) and window sizes (81, 163, and 244 samples, corresponding to 10 s, 20 s, and 30 s, respectively). The performance under the SI setup was notably lower than that observed with the random-split approach (Tables 3 and 4). This outcome is expected, as the SI evaluation represents a more challenging and realistic condition, in which the model must generalize to unseen participants. Such a setup minimizes the likelihood of data leakage and provides a more accurate reflection of real-world performance for cognitive load classification in driving scenarios.

Across all feature extraction methods, the accuracy values generally ranged between 0.44 and 0.53, with corresponding AUC values between approximately 0.55 and 0.69. These moderate values indicate that while the model captures some discriminative patterns related to cognitive load, there remains substantial inter-subject variability in fNIRS responses. Among the three feature extraction methods, ANOVA and FastICA consistently outperformed PCA in most cases, especially at a learning rate of 0.01, where both methods achieved relatively higher accuracies and more stable F1-scores across window sizes. The 20 s window (163 samples) often provided a better trade-off between performance and temporal resolution, suggesting that this duration may capture the most informative hemodynamic changes associated with cognitive load.

Table 7: Performance metrics for SI evaluation of the EEGNet model using overlapping window segmentation. Results are presented for ANOVA, PCA, and FastICA feature extraction methods across three learning rates (0.1, 0.01, 0.001) and window sizes (10 s, 20 s, and 30 s).

| Feature types | Learning rate | Window size | Accuracy (Mean ± SD) | AUC (Mean ± SD) | Recall (Mean ± SD) | Precision (Mean ± SD) | F1-Score (Mean ± SD) |
|---|---|---|---|---|---|---|---|
| ANOVA | 0.1 | 81 (10s) | 0.4683 ± 0.0199 | 0.6242 ± 0.0268 | 0.4683 ± 0.0199 | 0.4977 ± 0.1112 | 0.4977 ± 0.1112 |
| | | 163 (20s) | 0.4625 ± 0.0486 | 0.6186 ± 0.0794 | 0.4625 ± 0.0486 | 0.3854 ± 0.1016 | 0.3854 ± 0.1016 |
| | | 244 (30s) | 0.4585 ± 0.0490 | 0.6261 ± 0.0374 | 0.4585 ± 0.0490 | 0.4268 ± 0.0949 | 0.4268 ± 0.0949 |
| | 0.01 | 81 (10s) | 0.4902 ± 0.0334 | 0.6372 ± 0.0668 | 0.4902 ± 0.0334 | 0.5368 ± 0.0922 | 0.5368 ± 0.0922 |
| | | 163 (20s) | 0.5303 ± 0.0345 | 0.6802 ± 0.0506 | 0.5303 ± 0.0345 | 0.5681 ± 0.0626 | 0.5681 ± 0.0626 |
| | | 244 (30s) | 0.4983 ± 0.0515 | 0.6423 ± 0.0730 | 0.4983 ± 0.0515 | 0.4533 ± 0.1188 | 0.4533 ± 0.1188 |
| | 0.001 | 81 (10s) | 0.4861 ± 0.0404 | 0.6497 ± 0.0365 | 0.4861 ± 0.0404 | 0.5764 ± 0.0999 | 0.5764 ± 0.0999 |
| | | 163 (20s) | 0.5061 ± 0.0781 | 0.6548 ± 0.0864 | 0.5061 ± 0.0781 | 0.5400 ± 0.0890 | 0.5400 ± 0.0890 |
| | | 244 (30s) | 0.4787 ± 0.0402 | 0.6516 ± 0.0658 | 0.4787 ± 0.0402 | 0.4894 ± 0.0309 | 0.4894 ± 0.0309 |
| PCA | 0.1 | 81 (10s) | 0.4481 ± 0.0272 | 0.5566 ± 0.0806 | 0.4481 ± 0.0272 | 0.4215 ± 0.0801 | 0.4215 ± 0.0801 |
| | | 163 (20s) | 0.4643 ± 0.0288 | 0.6252 ± 0.0706 | 0.4643 ± 0.0288 | 0.5183 ± 0.1246 | 0.5183 ± 0.1246 |
| | | 244 (30s) | 0.4437 ± 0.0290 | 0.6496 ± 0.0666 | 0.4437 ± 0.0290 | 0.3924 ± 0.0833 | 0.3924 ± 0.0833 |
| | 0.01 | 81 (10s) | 0.4781 ± 0.0363 | 0.6346 ± 0.0707 | 0.4781 ± 0.0363 | 0.4272 ± 0.0509 | 0.4272 ± 0.0509 |
| | | 163 (20s) | 0.4986 ± 0.0498 | 0.6636 ± 0.0529 | 0.4986 ± 0.0498 | 0.4776 ± 0.1405 | 0.4776 ± 0.1405 |
| | | 244 (30s) | 0.4888 ± 0.0293 | 0.6529 ± 0.0588 | 0.4888 ± 0.0293 | 0.4423 ± 0.1432 | 0.4423 ± 0.1432 |
| | 0.001 | 81 (10s) | 0.4768 ± 0.0369 | 0.6256 ± 0.0360 | 0.4768 ± 0.0369 | 0.5080 ± 0.0588 | 0.5080 ± 0.0588 |
| | | 163 (20s) | 0.5155 ± 0.0697 | 0.6888 ± 0.0794 | 0.5155 ± 0.0697 | 0.5176 ± 0.1313 | 0.5176 ± 0.1313 |
| | | 244 (30s) | 0.4963 ± 0.0706 | 0.6576 ± 0.0760 | 0.4963 ± 0.0706 | 0.5295 ± 0.0952 | 0.5295 ± 0.0952 |
| FastICA | 0.1 | 81 (10s) | 0.4463 ± 0.0357 | 0.6203 ± 0.0368 | 0.4463 ± 0.0357 | 0.4673 ± 0.1735 | 0.4673 ± 0.1735 |
| | | 163 (20s) | 0.4438 ± 0.0231 | 0.5971 ± 0.0395 | 0.4438 ± 0.0231 | 0.4663 ± 0.0929 | 0.4663 ± 0.0929 |
| | | 244 (30s) | 0.4569 ± 0.0554 | 0.6303 ± 0.0343 | 0.4569 ± 0.0554 | 0.4174 ± 0.1447 | 0.4174 ± 0.1447 |
| | 0.01 | 81 (10s) | 0.4944 ± 0.0475 | 0.6415 ± 0.0924 | 0.4944 ± 0.0475 | 0.4195 ± 0.0896 | 0.4195 ± 0.0896 |
| | | 163 (20s) | 0.4820 ± 0.0433 | 0.6663 ± 0.0397 | 0.4820 ± 0.0433 | 0.4985 ± 0.1067 | 0.4985 ± 0.1067 |
| | | 244 (30s) | 0.4765 ± 0.0493 | 0.6585 ± 0.0571 | 0.4765 ± 0.0493 | 0.4881 ± 0.0987 | 0.4881 ± 0.0987 |
| | 0.001 | 81 (10s) | 0.4867 ± 0.0349 | 0.6635 ± 0.0408 | 0.4867 ± 0.0349 | 0.5003 ± 0.1053 | 0.5003 ± 0.1053 |
| | | 163 (20s) | 0.5046 ± 0.0702 | 0.6601 ± 0.0898 | 0.5046 ± 0.0702 | 0.5456 ± 0.0638 | 0.5456 ± 0.0638 |
| | | 244 (30s) | 0.4823 ± 0.0524 | 0.6627 ± 0.0547 | 0.4823 ± 0.0524 | 0.5118 ± 0.0549 | 0.5118 ± 0.0549 |

Table 8 presents the SI classification results for non-overlapping fNIRS signal segments using the EEGNet model across different feature extraction methods (ANOVA, PCA, and FastICA), learning rates, and window sizes. In contrast to the overlapping segmentation approach, non-overlapping windows result in slightly lower overall performance across all metrics. Accuracy values generally range between 0.48 and 0.56, with corresponding AUC values between 0.63 and 0.70, indicating moderate discriminative capability under cross-subject evaluation. Among the feature extraction methods, PCA and FastICA show marginally higher stability across varying learning rates, while ANOVA occasionally produces higher recall values, suggesting better sensitivity but at the cost of precision. Across the hyperparameter spectrum, a learning rate of 0.1 consistently produces better average accuracies compared to smaller learning rates (0.01 and 0.001). This suggests that faster convergence early in training aids the model in establishing more discriminative representations under limited subject-level variability. However, for smaller learning rates, the network tends to converge more slowly, which may reduce its ability to capture subtle yet subject-specific activation differences.

Table 8: SI classification performance of the EEGNet model using non-overlapping fNIRS signal segmentation across ANOVA, PCA, and FastICA feature extraction methods. Results are reported as mean ± standard deviation for Accuracy, AUC, Recall, Precision, and F1-score across three learning rates (0.1, 0.01, 0.001) and window sizes (10 s, 20 s, 30 s). The results demonstrate moderate generalization across unseen subjects, with performance influenced by window size, feature extraction technique, and learning rate.

| Feature types | Learning rate | Window size | Accuracy | AUC | Recall | Precision | F1-score |
|---|---|---|---|---|---|---|---|
| ANOVA | 0.1 | 81 (10s) | 0.5379 ± 0.0321 | 0.6738 ± 0.0631 | 0.5379 ± 0.0321 | 0.6462 ± 0.0487 | 0.6462 ± 0.0487 |
| | | 163 (20s) | 0.5484 ± 0.0214 | 0.6657 ± 0.0462 | 0.5484 ± 0.0214 | 0.6082 ± 0.0710 | 0.6082 ± 0.0710 |
| | | 244 (30s) | 0.5197 ± 0.0756 | 0.6563 ± 0.0826 | 0.5197 ± 0.0756 | 0.5519 ± 0.0659 | 0.5519 ± 0.0659 |
| | 0.01 | 81 (10s) | 0.4954 ± 0.0575 | 0.6536 ± 0.0633 | 0.4954 ± 0.0575 | 0.4926 ± 0.0949 | 0.4926 ± 0.0949 |
| | | 163 (20s) | 0.5213 ± 0.0757 | 0.6590 ± 0.1129 | 0.5213 ± 0.0757 | 0.5613 ± 0.0739 | 0.5613 ± 0.0739 |
| | | 244 (30s) | 0.4942 ± 0.0504 | 0.6418 ± 0.0820 | 0.4942 ± 0.0504 | 0.4825 ± 0.0834 | 0.4825 ± 0.0834 |
| | 0.001 | 81 (10s) | 0.5134 ± 0.0476 | 0.6528 ± 0.0509 | 0.5134 ± 0.0476 | 0.5294 ± 0.1113 | 0.5294 ± 0.1113 |
| | | 163 (20s) | 0.4921 ± 0.0549 | 0.6500 ± 0.0461 | 0.4921 ± 0.0549 | 0.5411 ± 0.0234 | 0.5411 ± 0.0234 |
| | | 244 (30s) | 0.5066 ± 0.0274 | 0.6305 ± 0.0651 | 0.5066 ± 0.0274 | 0.5565 ± 0.0390 | 0.5565 ± 0.0390 |
| PCA | 0.1 | 81 (10s) | 0.5261 ± 0.0655 | 0.6347 ± 0.0677 | 0.5261 ± 0.0655 | 0.5961 ± 0.0755 | 0.5961 ± 0.0755 |
| | | 163 (20s) | 0.5611 ± 0.0453 | 0.6974 ± 0.0542 | 0.5611 ± 0.0453 | 0.5744 ± 0.0847 | 0.5744 ± 0.0847 |
| | | 244 (30s) | 0.5111 ± 0.0354 | 0.6445 ± 0.0459 | 0.5111 ± 0.0354 | 0.5706 ± 0.0685 | 0.5706 ± 0.0685 |
| | 0.01 | 81 (10s) | 0.5041 ± 0.0607 | 0.6637 ± 0.0756 | 0.5041 ± 0.0607 | 0.5265 ± 0.1058 | 0.5265 ± 0.1058 |
| | | 163 (20s) | 0.5319 ± 0.0876 | 0.6340 ± 0.0813 | 0.5319 ± 0.0876 | 0.5706 ± 0.0820 | 0.5706 ± 0.0820 |
| | | 244 (30s) | 0.4859 ± 0.0618 | 0.6169 ± 0.0744 | 0.4859 ± 0.0618 | 0.5138 ± 0.0514 | 0.5138 ± 0.0514 |
| | 0.001 | 81 (10s) | 0.5130 ± 0.0469 | 0.6516 ± 0.0502 | 0.5130 ± 0.0469 | 0.5273 ± 0.1100 | 0.5273 ± 0.1100 |
| | | 163 (20s) | 0.4945 ± 0.0564 | 0.6526 ± 0.0484 | 0.4945 ± 0.0564 | 0.5433 ± 0.0222 | 0.5433 ± 0.0222 |
| | | 244 (30s) | 0.5107 ± 0.0264 | 0.6289 ± 0.0642 | 0.5107 ± 0.0264 | 0.5699 ± 0.0443 | 0.5699 ± 0.0443 |
| FastICA | 0.1 | 81 (10s) | 0.5454 ± 0.0496 | 0.6748 ± 0.0854 | 0.5454 ± 0.0496 | 0.5948 ± 0.0375 | 0.5948 ± 0.0375 |
| | | 163 (20s) | 0.5422 ± 0.0628 | 0.6896 ± 0.0784 | 0.5422 ± 0.0628 | 0.6123 ± 0.0722 | 0.6123 ± 0.0722 |
| | | 244 (30s) | 0.5162 ± 0.0751 | 0.6549 ± 0.0996 | 0.5162 ± 0.0751 | 0.5631 ± 0.0779 | 0.5631 ± 0.0779 |
| | 0.01 | 81 (10s) | 0.4941 ± 0.0688 | 0.6613 ± 0.0502 | 0.4941 ± 0.0688 | 0.4622 ± 0.1181 | 0.4622 ± 0.1181 |
| | | 163 (20s) | 0.5342 ± 0.0878 | 0.6381 ± 0.0827 | 0.5342 ± 0.0878 | 0.5739 ± 0.0837 | 0.5739 ± 0.0837 |
| | | 244 (30s) | 0.4871 ± 0.0452 | 0.6354 ± 0.0638 | 0.4871 ± 0.0452 | 0.5129 ± 0.0345 | 0.5129 ± 0.0345 |
| | 0.001 | 81 (10s) | 0.5138 ± 0.0478 | 0.6519 ± 0.0502 | 0.5138 ± 0.0478 | 0.5275 ± 0.1104 | 0.5275 ± 0.1104 |
| | | 163 (20s) | 0.4945 ± 0.0571 | 0.6524 ± 0.0485 | 0.4945 ± 0.0571 | 0.5436 ± 0.0243 | 0.5436 ± 0.0243 |
| | | 244 (30s) | 0.5102 ± 0.0258 | 0.6276 ± 0.0664 | 0.5102 ± 0.0258 | 0.5596 ± 0.0373 | 0.5596 ± 0.0373 |

In terms of window size, shorter windows (10s and 20s) tend to outperform longer ones (30s), aligning with the hypothesis that shorter temporal segments capture transient cognitive fluctuations more effectively in SI scenarios. However, the overall performance reduction relative to overlapping segmentation highlights the importance of temporal continuity in fNIRS-based cognitive load detection. The loss of temporal overlap between adjacent windows may limit the model's ability to capture smooth transitions in neural activity patterns, leading to lower generalization across unseen participants.

In this study, we further extended the SI evaluation by incorporating an automatic learning rate adjustment strategy, specifically the RLRP method, to optimize the convergence behavior of the EEGNet model. This technique dynamically reduces the learning rate when the validation loss stops improving after a predefined number of epochs (patience = 2), thereby preventing overfitting and allowing the optimizer to converge more effectively toward an optimal solution. Figure 14a, Figure 14b, and Figure 14c depict the SI training behaviour of EEGNet using ANOVA, PCA, and FastICA features, respectively, under both overlapping and non-overlapping segmentation strategies. These plots show how the model learns under fixed learning rates when trained across participants, a scenario where greater signal variability and inter-subject differences markedly increase optimization difficulty.

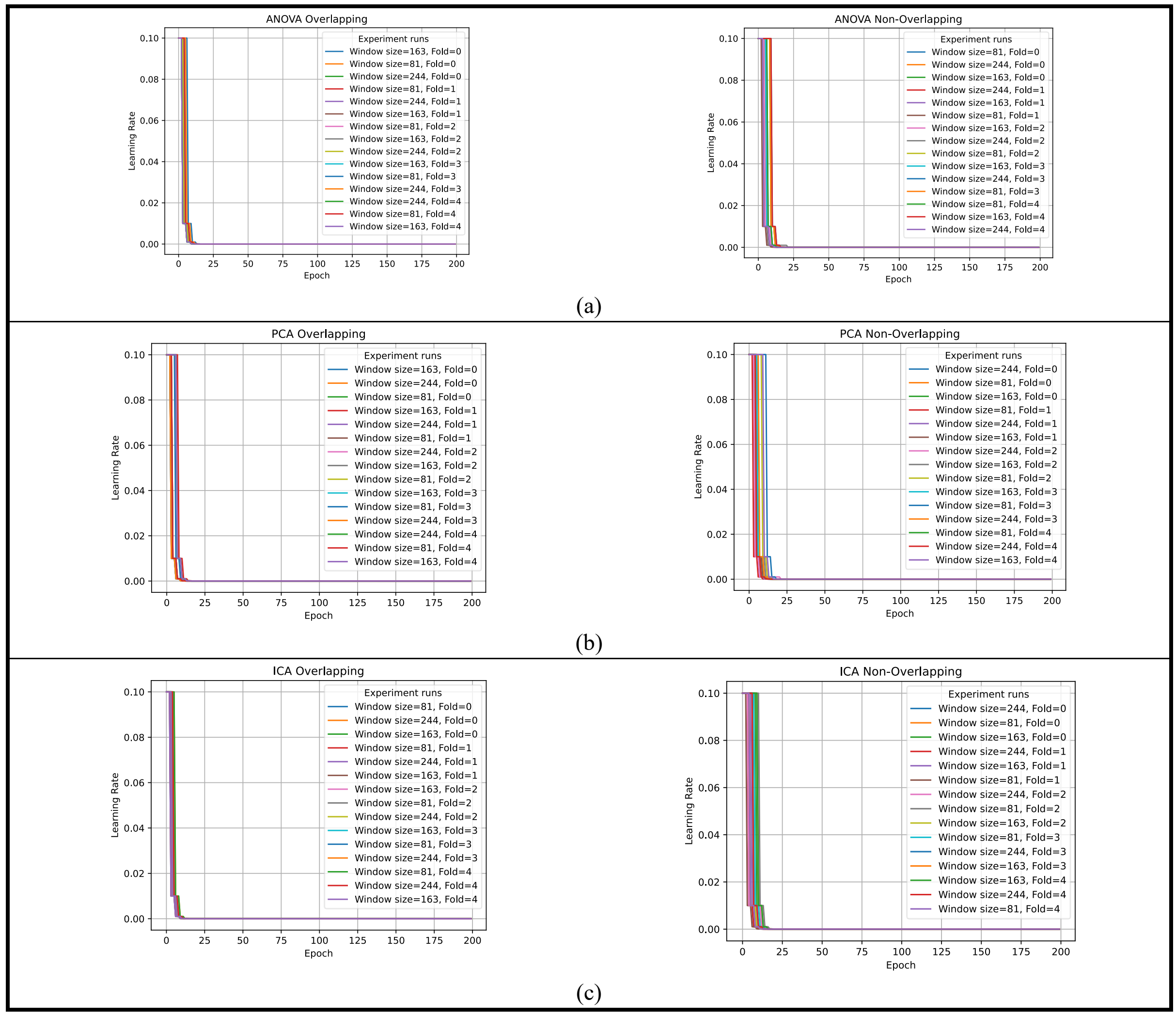

Figure 14: Subject-independent training behaviour of the EEGNet model using (a) ANOVA, (b) PCA, and (c) FastICA features under overlapping and non-overlapping temporal segmentation. The plots illustrate convergence patterns and highlight the increased optimization difficulty arising from inter-subject variability in fNIRS signals.

Table 9 presents the model's performance when applying the RLRP strategy to overlapping fNIRS signal segments. The results demonstrate stable yet moderate classification performance across all three feature extraction methods (ANOVA, PCA, and FastICA) and window sizes (10 s, 20 s, and 30 s). Accuracy values range between 0.42 and 0.48, and AUC values between 0.60 and 0.67, which are consistent with previous SI results (Table 8) but show slightly improved stability (lower standard deviation) across folds. This suggests that automatic learning rate adaptation effectively mitigates fluctuations in optimization, promoting more balanced learning across subjects. Among the feature extraction methods, FastICA exhibits the highest average accuracy (0.4828 ± 0.0400) and best recall performance, indicating its advantage in capturing spatially independent and temporally stable features across individuals. PCA also shows competitive results, maintaining moderate recall and precision balance, while ANOVA tends to perform inconsistently, with relatively lower AUC values at smaller window sizes (e.g., 0.5867 ± 0.0580 for 20 s). Regarding window size effects, longer segments (30 s, 244 samples) yield marginally higher AUC scores across all methods, implying that extended temporal context improves discrimination when overlapping windows are used. The presence of overlapping time windows enhances the model's ability to utilize fine-grained temporal dependencies between adjacent samples, thereby reinforcing the effect of adaptive learning rate scheduling.

Table 9: SI EEGNet performance using overlapping fNIRS signal segmentation with automatic learning rate adaptation (Reduce Learning Rate on Plateau). Results are shown across 5 folds for ANOVA, PCA, and FastICA feature extraction methods, with window sizes of 10 s, 20 s, and 30 s, respectively.

| Sample size | Feature extraction | Accuracy (Mean ± SD) | AUC (Mean ± SD) | Recall (Mean ± SD) | Precision (Mean ± SD) | F1-Score (Mean ± SD) |
|---|---|---|---|---|---|---|

| | methods | | | | | |
|---|---|---|---|---|---|---|
| **81 (10s)** | ANOVA | 0.4662 ± 0.0437 | 0.6194 ± 0.0785 | 0.4662 ± 0.0437 | 0.4242 ± 0.1110 | 0.4242 ± 0.1110 |
| | PCA | 0.4679 ± 0.0356 | 0.6188 ± 0.0737 | 0.4679 ± 0.0356 | 0.5069 ± 0.0583 | 0.5069 ± 0.0583 |
| | FastICA | 0.4770 ± 0.0533 | 0.6097 ± 0.0856 | 0.4770 ± 0.0533 | 0.5945 ± 0.0673 | 0.5945 ± 0.0673 |
| **163 (20s)** | ANOVA | 0.4288 ± 0.0185 | 0.5867 ± 0.0580 | 0.4288 ± 0.0185 | 0.4418 ± 0.0338 | 0.4418 ± 0.0338 |
| | PCA | 0.4611 ± 0.0526 | 0.6421 ± 0.0465 | 0.4611 ± 0.0526 | 0.4640 ± 0.0739 | 0.4640 ± 0.0739 |
| | FastICA | 0.4722 ± 0.0181 | 0.6324 ± 0.0586 | 0.4722 ± 0.0181 | 0.5098 ± 0.0437 | 0.5098 ± 0.0437 |
| **244 (30s)** | ANOVA | 0.4684 ± 0.0252 | 0.6669 ± 0.0520 | 0.4684 ± 0.0252 | 0.4921 ± 0.0431 | 0.4921 ± 0.0431 |
| | PCA | 0.4655 ± 0.0363 | 0.6413 ± 0.0539 | 0.4655 ± 0.0363 | 0.5272 ± 0.0434 | 0.5272 ± 0.0434 |
| | FastICA | 0.4828 ± 0.0400 | 0.6356 ± 0.0418 | 0.4828 ± 0.0400 | 0.4924 ± 0.0528 | 0.4924 ± 0.0528 |

Table 10 reports the corresponding results for non-overlapping segmentation, where each segment is temporally independent. Across all configurations, the model maintains accuracy between 0.43 and 0.49 and AUC between 0.59 and 0.65, which are slightly lower than those obtained for overlapping windows. However, the inclusion of automatic learning rate scheduling clearly stabilizes the training dynamics, reflected in smaller variability across folds (standard deviations typically below 0.05). PCA-based features demonstrate the most consistent performance in this setting, achieving the highest mean AUC (0.6536 ± 0.0544) and balanced precision-recall trade-offs. FastICA follows closely, indicating that both dimensionality-reduction methods generalize better than ANOVA in cross-subject conditions. Conversely, ANOVA features exhibit more fluctuation, possibly due to their dependence on within-subject variance, which does not translate well across unseen participants. In contrast to the overlapping setup, shorter window sizes (10 s and 20 s) yield slightly higher accuracy and recall than longer ones (30 s).

Table 10: SI EEGNet performance using non-overlapping fNIRS signal segmentation with automatic learning rate adaptation (Reduce Learning Rate on Plateau). Results are shown across 5 folds for ANOVA, PCA, and FastICA feature extraction methods, with window sizes of 10 s, 20 s, and 30 s, respectively.

| **Sample size** | **Feature extraction methods** | **Accuracy (Mean ± SD)** | **AUC (Mean ± SD)** | **Recall (Mean ± SD)** | **Precision (Mean ± SD)** | **F1-Score (Mean ± SD)** |
|---|---|---|---|---|---|---|
| **81 (10s)** | ANOVA | 0.4577 ± 0.0100 | 0.6171 ± 0.0374 | 0.4577 ± 0.0100 | 0.4928 ± 0.0626 | 0.4928 ± 0.0626 |
| | PCA | 0.4874 ± 0.0399 | 0.6536 ± 0.0544 | 0.4874 ± 0.0399 | 0.5227 ± 0.0748 | 0.5227 ± 0.0748 |
| | FastICA | 0.4655 ± 0.0156 | 0.6208 ± 0.0457 | 0.4655 ± 0.0156 | 0.5119 ± 0.0520 | 0.5119 ± 0.0520 |
| **163 (20s)** | ANOVA | 0.4637 ± 0.0342 | 0.6317 ± 0.0730 | 0.4637 ± 0.0342 | 0.5379 ± 0.0686 | 0.5379 ± 0.0686 |
| | PCA | 0.4775 ± 0.0267 | 0.6363 ± 0.0520 | 0.4775 ± 0.0267 | 0.5131 ± 0.0476 | 0.5131 ± 0.0476 |
| | FastICA | 0.4644 ± 0.0362 | 0.6405 ± 0.0653 | 0.4644 ± 0.0362 | 0.4884 ± 0.0317 | 0.4884 ± 0.0317 |
| **244 (30s)** | ANOVA | 0.4524 ± 0.0310 | 0.6208 ± 0.1202 | 0.4524 ± 0.0310 | 0.4618 ± 0.0278 | 0.4618 ± 0.0278 |
| | PCA | 0.4338 ± 0.0300 | 0.5944 ± 0.0994 | 0.4338 ± 0.0300 | 0.4711 ± 0.0216 | 0.4711 ± 0.0216 |
| | FastICA | 0.4284 ± 0.0410 | 0.6001 ± 0.0975 | 0.4284 ± 0.0410 | 0.4558 ± 0.0397 | 0.4558 ± 0.0397 |

From the results, it is evident that the fixed learning rate strategy consistently outperforms the automatic learning rate adaptation across all experimental configurations. This suggests that manual control of the learning rate provides a more stable and effective optimization path for the EEGNet model in the context of SI fNIRS-based cognitive load classification. The model performance using the fixed learning rate approach across the three feature extraction methods ANOVA (Figure 15a), PCA (Figure 15b), and FastICA (Figure 15c)was evaluated at learning rates of 0.1, 0.01, and 0.001, and across window sizes of 10, 20, and 30 seconds, under both overlapping and non-overlapping temporal segmentation conditions. The 3D surface plots clearly depict the interactive effects of learning rate, temporal window length, and feature type on model performance metrics such as accuracy and AUC. A general trend observed across all configurations indicates that performance improved as the learning rate decreased, implying that smaller learning rates enabled more gradual and stable convergence, reducing oscillations during optimization and mitigating potential overfitting. This effect was particularly pronounced in the SI setup, where inter-subject variability introduces additional learning challenges. Furthermore, larger window sizes (20–30 seconds) consistently yielded more reliable and smoother performance patterns, reinforcing the notion that longer temporal segments capture richer contextual and physiological information, thereby enhancing the model's discriminative ability to classify varying cognitive load levels.

When analyzing results by feature extraction method, ANOVA-based features achieved the highest accuracy under overlapping window segmentation with a learning rate of 0.1 and a 10-second window. This observation emphasizes the benefit of increased temporal continuity and denser data sampling, which strengthens the reliability of statistical feature computation in ANOVA. However, in the non-overlapping configuration, ANOVA features demonstrated more balanced but slightly lower performance, with the best results emerging at a learning rate of 0.001, indicating improved generalization under slower learning conditions. PCA-based features exhibited stable performance across both segmentation strategies, showing the best overall accuracy with non-overlapping windows, particularly at a 20-second window and a learning rate of 0.1. This suggests that PCA effectively captures dominant patterns of variance within the fNIRS signals, even without relying on temporal overlap, thus maintaining high generalization capability. In contrast, FastICA-based features demonstrated the most consistent and robust behavior across all experimental settings. The highest performance was obtained under non-overlapping conditions, with a 20-second window and a learning rate of 0.1, indicating that FastICA isolates statistically independent neural components that generalize well across participants.

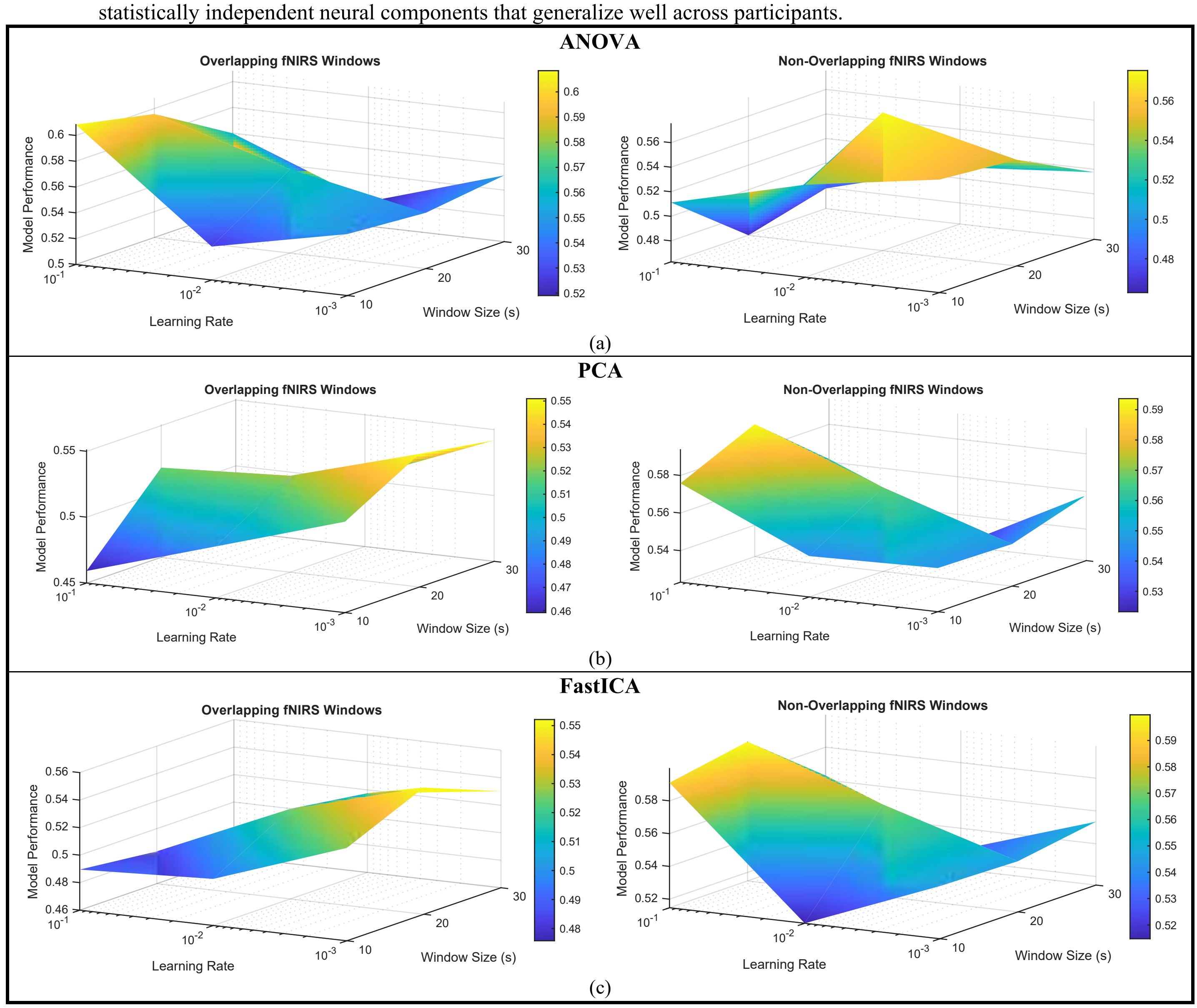

Figure 15: Model performance using the fixed learning rate approach across three feature extraction methods: (a) ANOVA, (b) PCA, and (c) FastICA. Each surface plot illustrates performance variations across window sizes of 10, 20, and 30 seconds and learning rates of 0.1, 0.01, and 0.001, evaluated under both overlapping and non-overlapping temporal segmentation conditions.

To investigate the impact of segmentation strategy on EEGNet performance, we conducted a comparative analysis between overlapping and non-overlapping segments across three feature extraction techniques: ANOVA, PCA, and FastICA. The evaluation employed standard performance metrics, including accuracy, precision, recall, F1-score, and area under the ROC curve (AUC), allowing a comprehensive assessment of model efficacy. In all cases, negative values indicate that the non-overlapping segmentation outperforms overlapping segmentation, providing a clear measure of the performance advantage. This analysis highlights how segmentation strategy interacts with hyperparameters such as window size and learning rate and examines the resulting implications for model generalization and robustness.

Figure 16a presents the performance differences of EEGNet trained with overlapping versus non-overlapping segmentation using ANOVA features. Across the majority of hyperparameter configurations, non-overlapping segmentation consistently outperforms overlapping segmentation, demonstrating that redundancy introduced by overlapping segments reduces model efficacy. The performance advantage is most pronounced at the smallest window size of 10 seconds, where non-overlapping segmentation improves accuracy by up to 14.9% at a learning rate of 0.1 and shows similarly substantial gains at lower learning rates. At a 20-second window, some hyperparameter settings show near parity, with the overlapping method achieving a marginal improvement of -1.0%, but overall, non-overlapping segmentation remains the superior approach. These results indicate that ANOVA features benefit from a less correlated and more diverse training set, which non-overlapping segmentation provides, leading to improved generalization and model reliability.

Figure 16b shows EEGNet performance differences using PCA features with overlapping versus non-overlapping segmentation. The trend mirrors that observed with ANOVA features, with non-overlapping segmentation consistently yielding higher accuracy across nearly all window sizes and learning rates. The most substantial performance deficits for overlapping segments occur at a 10-second window, with improvements of -14.9%, -18.6%, and -13.3% across learning rates of 0.1, 0.01, and 0.001, respectively. At a 20-second window, one configuration exhibits near parity, but non-overlapping segmentation regains the advantage for other learning rates. These findings suggest that even when using dimensionality reduction, overlapping segments introduce redundancy that negatively impacts model performance. Consequently, non-overlapping segmentation provides a more robust and less correlated training dataset, improving EEGNet generalization when using PCA features.

Figure 16c depicts EEGNet performance differences with FastICA features. Here, the benefits of non-overlapping segmentation are even more pronounced, particularly at the 10-second window, where accuracy improvements reach 22.2%, 22.2%, and 13.0% for learning rates of 0.1, 0.01, and 0.001, respectively. FastICA inherently aims to extract statistically independent components, and non-overlapping segmentation preserves this independence by avoiding redundant training samples, resulting in dramatic performance gains. At larger window sizes (20–30 seconds), the performance gap decreases but remains decisively in favor of non-overlapping segmentation, with only a single near-parity instance (-0.1%) observed. Overall, non-overlapping segmentation is critical for maximizing EEGNet performance with FastICA features, providing a diverse, less correlated training dataset that enhances model robustness and generalization.

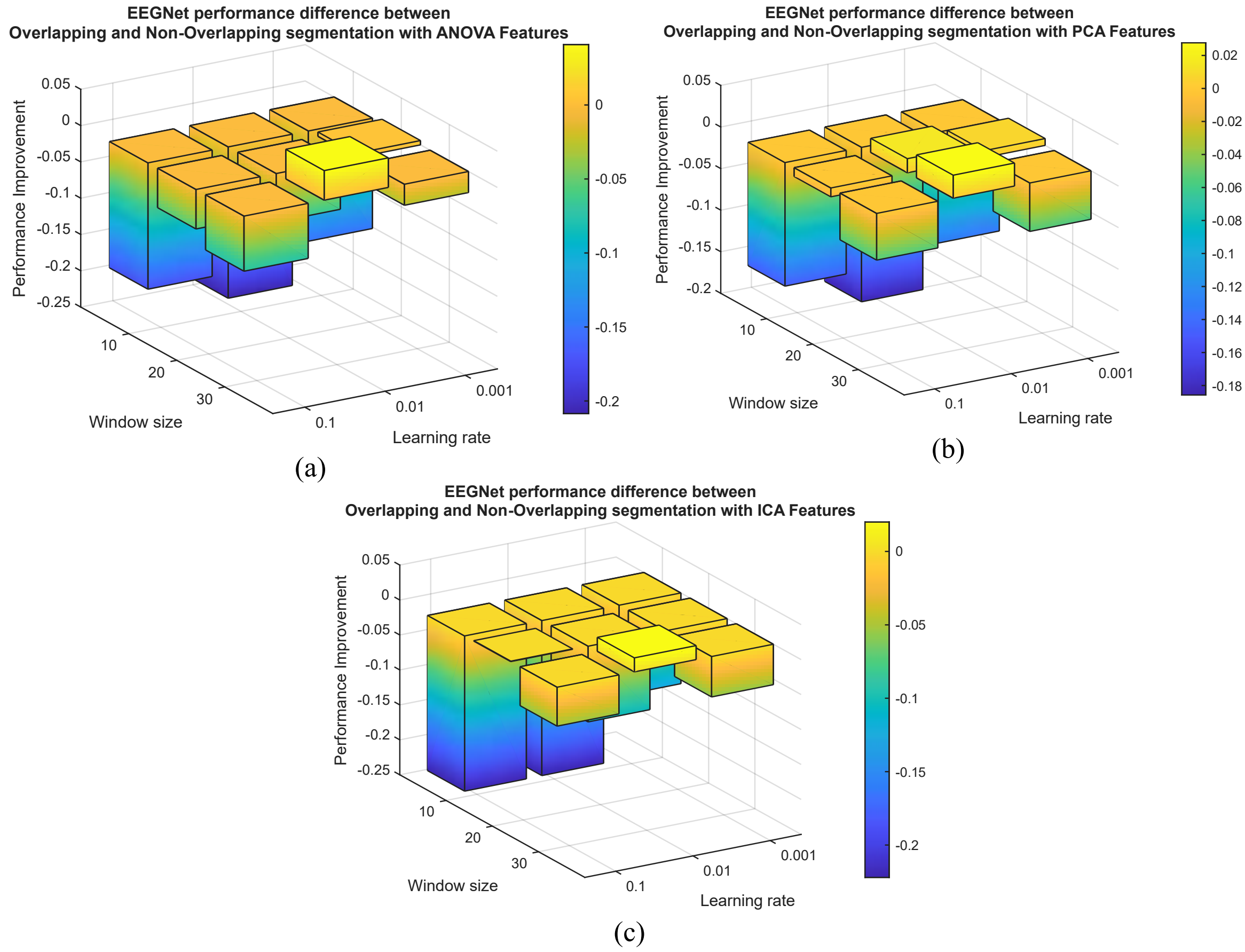

Figure 16: Comparative analysis of EEGNet performance using overlapping versus non-overlapping fNIRS signal segments across three feature extraction methods. (a) ANOVA features, (b) PCA features, and (c) FastICA features. Negative values indicate performance gains from non-overlapping segmentation. Results highlight how non-overlapping segments improve model generalization, particularly at smaller window sizes and across different learning rates.

### *4.5. Comparative evaluation of EEGNet on fNIRS data*

The results of this study highlight several key factors influencing EEGNet's performance in fNIRS-based cognitive load classification, namely temporal segmentation, learning rate, feature extraction method, and evaluation protocol. As illustrated in Figure 17, which presents the performance heatmaps for both random and SI data splits, temporal segmentation emerged as a dominant factor. Overlapping windows consistently yielded the highest performance metrics across all window sizes and learning rates, particularly at lower learning rates (0.01 and 0.001), where accuracy approached near-perfect levels. In contrast, non-overlapping windows achieved lower accuracies (83-97%) but offered a more realistic estimate of generalization, with shorter windows (10 s, 81 samples) outperforming longer ones (30 s, 244 samples). Learning rate selection was also critical. Smaller fixed rates (e.g., 0.001) promoted stable convergence and improved accuracy, whereas larger rates (e.g., 0.1) frequently caused premature convergence or oscillations, especially in non-overlapping configurations. The superior performance observed under overlapping segmentation can likely be attributed to temporal redundancy, since random splits, even under a 5-fold scheme, may share overlapping data segments. This effectively increases the density of training samples and allows the model to capture fine-grained hemodynamic transitions. However, such overlap may also introduce data leakage, potentially leading to an overestimation of generalization performance.

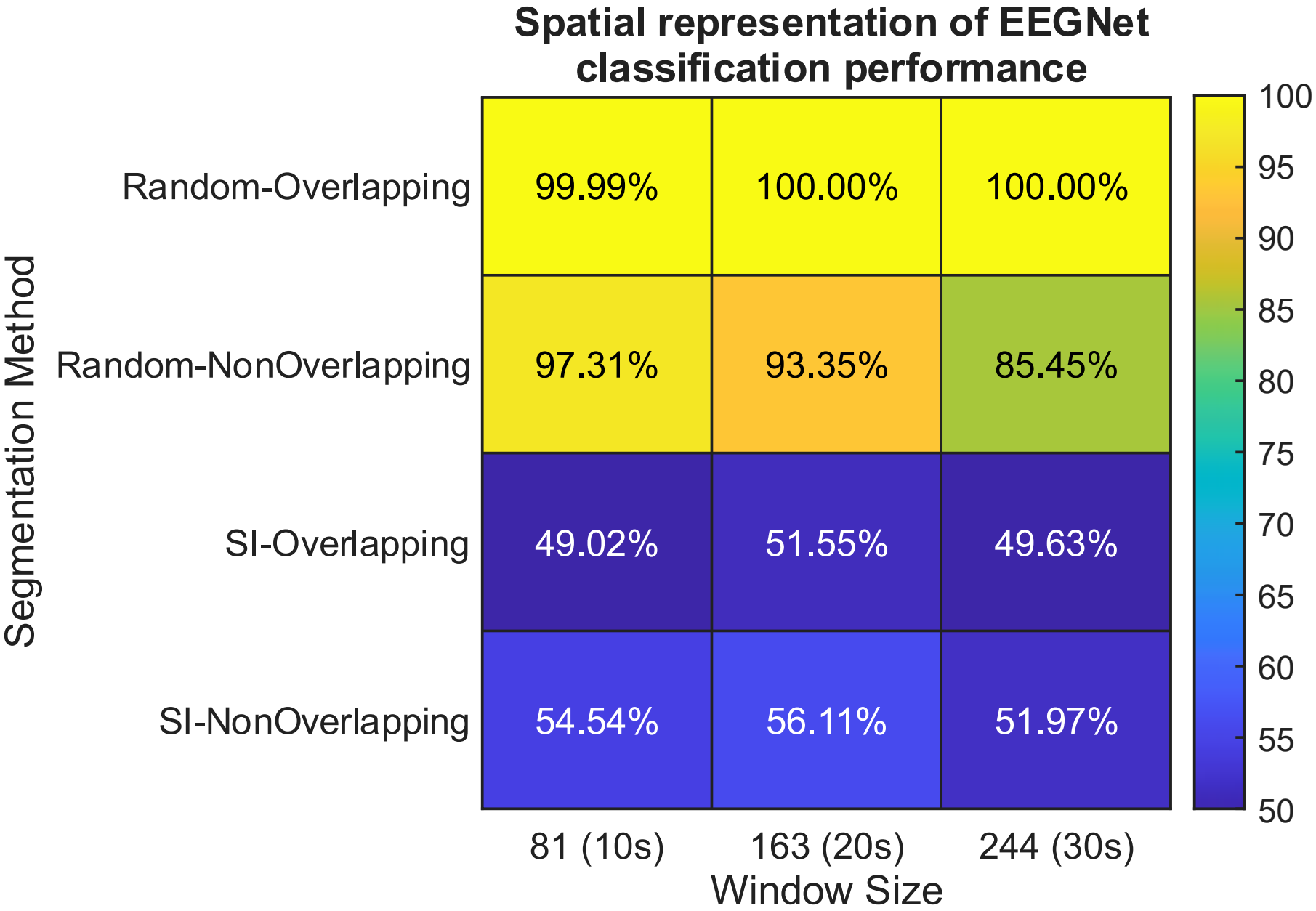

Figure 17: Performance heat maps of EEGNet for random and SI data splits across different temporal segmentation strategies, window sizes, and learning rates. The plots illustrate the influence of overlapping versus non-overlapping segmentation and feature extraction methods (ANOVA, PCA, FastICA) on model accuracy and generalization.

While random-split results demonstrate strong performance across various configurations, the SI results reveal a clear performance gap, indicating a reduction in generalization when the model is tested on unseen participants. This contrast underscores a central challenge in developing real-time, user-independent cognitive load classification systems using fNIRS data. For SI overlapping segmentation, performance remained moderate across all window sizes and feature extraction methods. The highest accuracy (49%) was achieved using ANOVA features with a 10-second window (81 samples) and a learning rate of 0.01. Increasing the window duration to 20 seconds (163 samples) and 30 seconds (244 samples) produced only marginal improvements, with PCA features yielding accuracies of 51.55% and 49.63%, respectively. These results suggest that longer temporal windows can capture more stable and contextually rich hemodynamic patterns, though inter-subject variability in fNIRS signals continues to limit overall generalization. Moreover, the modest performance gains from overlapping segmentation under SI evaluation imply that redundancy among samples may not be beneficial when training and testing data originate from different individuals, as it may amplify subject-specific correlations rather than generalizable neural patterns.

In contrast, non-overlapping segmentation achieved comparatively higher accuracies, suggesting its advantage in producing independent and diverse training instances that promote cross-subject generalization. The best SI performance was observed with PCA features using a 20-second window and a learning rate of 0.1, achieving an accuracy of 56.11%. FastICA features with a 10-second window and ANOVA features with a 30-second window also yielded competitive outcomes. These findings indicate that non-overlapping segmentation better captures SI signal representations, as eliminating temporal redundancy allows the model to learn more generalizable features of cognitive load, less biased by individual hemodynamic variability.

Finally, the RLRP adaptive learning rate strategy significantly improved the stability and convergence behavior of EEGNet during training for both random and SI evaluations. By automatically decreasing the learning rate when validation performance plateaued, RLRP effectively reduced oscillations and prevented premature convergence, issues frequently observed in deep learning models trained on physiological signals such as fNIRS. However, despite its stabilizing benefits, the RLRP approach did not surpass the performance of models trained with manually tuned fixed learning rates, particularly in terms of peak accuracy and generalization. Fixed learning rates, when optimally configured (e.g., 0.01 or 0.001), produced more consistent and superior classification outcomes. This suggests that EEGNet's learning dynamics in this context are better governed by predefined learning rate schedules rather than adaptive decay mechanisms.

## 5. Conclusion and future work

This study presented a comprehensive evaluation of the EEGNet architecture for cognitive workload classification using fNIRS signals. Through systematic experimentation with temporal segmentation strategies, feature extraction methods, and learning rate configurations, we identified several key factors influencing model performance and generalization. The results demonstrated that overlapping temporal windows and smaller fixed learning rates (0.001-0.01) yielded superior performance under random data splits, achieving near-perfect accuracy in some configurations. These findings suggest that temporal redundancy and fine-grained hemodynamic sampling enable EEGNet to effectively capture localized and transient cognitive load variations.

However, when evaluated under SI conditions, performance substantially declined, underscoring the challenges of inter-subject variability and the difficulty of achieving cross-participant generalization in fNIRS-based deep learning models. Non-overlapping segmentation outperformed overlapping windows in SI evaluation, indicating that reducing temporal redundancy helps the model learn more generalizable patterns rather than subject-specific signatures. This contrast between random and SI results highlights the importance of rigorous evaluation protocols and the need to interpret high within-subject accuracy with caution, as random data splits may introduce hidden data leakage. Moreover, while the RLRP adaptive strategy enhanced convergence stability, it did not outperform manually optimized fixed learning rates in terms of peak accuracy or robustness. These findings suggest that EEGNet's dynamics in this context benefit more from careful manual tuning than from adaptive decay mechanisms. The results emphasize that reliable fNIRS-based cognitive load classification requires not only an appropriate deep learning architecture but also a carefully designed data segmentation strategy and evaluation framework.

Future work will focus on enhancing the generalization capability of EEGNet for fNIRS-based cognitive load classification. This includes exploring domain adaptation techniques to mitigate inter-subject variability, as well as personalized calibration approaches that tailor the model to individual neural profiles without extensive retraining. Additionally, integrating multimodal physiological measures, such as ECG and eye-tracking data, will be pursued to capture complementary aspects of cognitive load, thereby improving robustness and transferability. We also plan to employ a driving simulator with a more advanced motion cueing algorithm to better replicate real-world driving dynamics, providing richer and more ecologically valid data for model training. By combining these strategies, we aim to develop a comprehensive framework capable of reliable, real-time assessment of driver cognitive states across diverse individuals and driving conditions, facilitating safer and more adaptive human–machine interactions.